\documentclass[journal]{IEEEtran}   
\ifCLASSINFOpdf
\else
\fi
\usepackage[table]{xcolor}
\usepackage{hhline} 
\usepackage{balance}  
\usepackage{multirow}
\usepackage{graphics}
\usepackage{times}
\usepackage{amsmath}
\usepackage{wasysym}  
\usepackage{wrapfig}
\usepackage{multirow}
\usepackage{amsthm}
\usepackage{amsfonts}
\usepackage{leftidx}
\usepackage{indentfirst}
\usepackage{lipsum}
\usepackage{epsfig}
\usepackage{prettyref}
\usepackage[e]{esvect} 
\usepackage{amssymb} 
\usepackage{graphicx}
\usepackage{algorithm}
\usepackage{algpseudocode}

\theoremstyle{plain}
\theoremstyle{definition}
\usepackage{enumerate}

\usepackage{hyperref}
\usepackage[sort,compress]{cite}

\usepackage{cases}

\makeatletter
\newcommand*{\algrule}[1][\algorithmicindent]{\makebox[#1][l]{\hspace*{.5em}\vrule height .75\baselineskip depth .25\baselineskip}}%

\newcount\ALG@printindent@tempcnta
\def\ALG@printindent{%
    \ifnum \theALG@nested>0
        \ifx\ALG@text\ALG@x@notext
            \addvspace{-3pt}
        \else
            \unskip
            \ALG@printindent@tempcnta=1
            \loop
                \algrule[\csname ALG@ind@\the\ALG@printindent@tempcnta\endcsname]%
                \advance \ALG@printindent@tempcnta 1
            \ifnum \ALG@printindent@tempcnta<\numexpr\theALG@nested+1\relax
            \repeat
        \fi
    \fi
    }%
\usepackage{etoolbox}
\patchcmd{\ALG@doentity}{\noindent\hskip\ALG@tlm}{\ALG@printindent}{}{\errmessage{failed to patch}}
\makeatother

\usepackage[framed,numbered,autolinebreaks,useliterate]{mcode}
\usepackage{booktabs}
\usepackage{tikz}
\usepackage{mathrsfs}  

\newcommand{\realfield}[1]{\hbox{I \kern -.5em R}^{#1}}
\newcommand {\mb}[1]{\mathbf{#1}} 
\newcommand {\bs}[1]{\boldsymbol{#1}}

\newcommand{\T}{^{\mathrm{T}}}  
\newcommand{\Rot}[2]{{^{#1}\mathbf{R}}_{#2}}  
\newcommand*\circled[1]
{\tikz[baseline=(char.base)]{\node[circle,minimum size=7pt,draw=black,inner sep=0.5pt](char){\scriptsize #1};}}

\usepackage{mathtools}
\usepackage{soul}

\usepackage{gensymb} 

\definecolor{orangeEA}{RGB}{217,83,25}

\usepackage{pifont}  

\usepackage{enumerate}

\newenvironment{packed_itemize}
{\begin{itemize}
  \setlength{\itemsep}{1pt}
  \setlength{\parskip}{0pt}
  \setlength{\parsep}{0pt}
}{\end{itemize}}

\usepackage{multirow}  
\usepackage{tabularx}  
\usepackage{colortbl} 
\usepackage{footnote}
\makeatletter
\newcommand{\thickhline}{%
    \noalign {\ifnum 0=`}\fi \hrule height 1pt
    \futurelet \reserved@a \@xhline
}
\newcolumntype{"}{@{\hskip\tabcolsep\vrule width 2pt\hskip\tabcolsep}}
\makeatother
\begin{document}
%
%
\title{Lie Group Formulation and Sensitivity Analysis for Shape Sensing of Variable Curvature Continuum Robots with General String Encoder Routing}
%
%
%

\author{Andrew~L.~Orekhov\textsuperscript{1,2},~Elan~Z.~Ahronovich\textsuperscript{1},~and~Nabil~Simaan\textsuperscript{1}
\thanks{\textsuperscript{1}Department of Mechanical Engineering, Vanderbilt University, Nashville, TN, USA. e-mail: {\tt aorekhov@andrew.cmu.edu, (elan.z.ahronovich, nabil.simaan)@vanderbilt.edu}}
\thanks{\textsuperscript{2}Robotics Institute, Carnegie Mellon University, Pittsburgh, PA, USA}
\thanks{This work was supported by NSF award \#CMMI-1734461 and by Vanderbilt University internal funds. A. Orekhov was partially supported by an NSF Graduate Research Fellowship under \#DGE-1445197.}
}

%
%

\markboth{\scriptsize \copyright 2022 IEEE. Accepted for Publication in IEEE Transactions on Robotics.}%
{Shell \MakeLowercase{\textit{et al.}}: Bare Demo of IEEEtran.cls for Journals}
%



\makeatletter
\let\NAT@parse\undefined
\makeatother
\maketitle
\begin{abstract}
This paper considers a combination of actuation tendons and measurement strings to achieve accurate shape sensing and direct kinematics of continuum robots.  Assuming general string routing, a methodical Lie group formulation for the shape sensing of these robots is presented. The shape kinematics is expressed using arc-length-dependent curvature distributions parameterized by modal functions, and the Magnus expansion for Lie group integration is used to express the shape as a product of exponentials. The tendon and string length kinematic constraints are solved for the modal coefficients and the configuration space and body Jacobian are derived. The noise amplification index for the shape reconstruction problem is defined and used for optimizing the string/tendon routing paths, and a planar simulation study shows the minimal number of strings/tendons needed for accurate shape reconstruction. A torsionally stiff continuum segment is used for experimental evaluation, demonstrating mean (maximal) end-effector absolute position error of less than $2\%$ ($5\%$) of total length. Finally, a simulation study of a torsionally compliant segment demonstrates the approach for general deflections and string routings. We believe that the methods of this paper can benefit the design process, sensing and control of continuum and soft robots.
\end{abstract}
\begin{IEEEkeywords}
Continuum robots, soft robots, shape sensing, Lie group methods, human-robot collaboration
\end{IEEEkeywords}
\section{Introduction}
\begin{figure}[htbp]
  \centering
  \includegraphics[width=0.95\columnwidth]{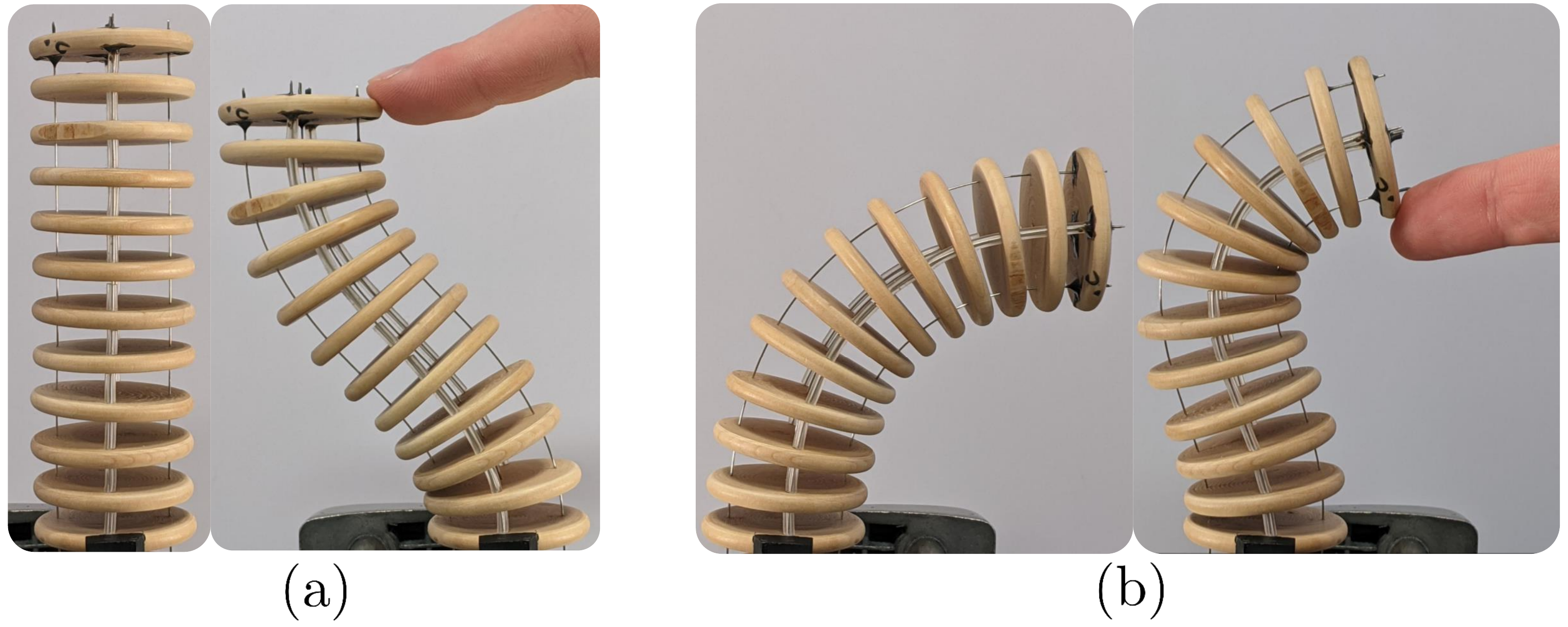}
  \caption{A continuum segment subject to passive deflections (a) starting in a straight configuration (b) starting with a bent configuration. This paper addresses the problem of sensing the deflected shape using string encoders.}
  \label{fig:deflection_examples}
\end{figure}

\IEEEPARstart{C}{ontinuum} robots achieve infinitely varied shapes governed by their elastic equilibrium conformations rather than being solely determined by the values of their active joints, as shown in Fig. \ref{fig:deflection_examples}. As a result, the direct kinematics of these robots is corrupted with a large level of uncertainty due to structural deflections. Since active joint values alone are not sufficient for achieving accurate direct kinematics, an additional sensing source is needed.  This need motivated efforts on shape sensing for continuum and soft robots, which focused on using extrinsic and/or intrinsic sensing.
\par Extrinsic sensing relies on use of external metrology (e.g. magnetic trackers \cite{SongRen2015_magntic_shape_continuum} and computer vision \cite{Camarillo2008vision,Reiter2011_shape_vision_continuum,Reiter2012_shape_vision_continuum,Reilink2013_vision_sensing_continuum}). Intrinsic sensing relies on sensors that augment joint-level data to produce shape estimates. The use of PVDF bimorph sensors for shape sensing of soft actuators under planar bending  was explored in \cite{ShapiroWolf2014_PVDF_shape_Sensing} along with considerations for the number of sensors and their spacing to avoid Runge's phenomenon when fitting a modal function series to estimate the shape. Others focused heavily on the use of fiber Bragg grating (FBG) sensors for shape sensing \cite{Cutkosky2010fbg,Patel2016fbg,Misra2014fbg_tmech}. A mechanics model combined with moment sensing was presented in \cite{ClementsRhan2006_whisker} to estimate the shape of whiskers and in \cite{trivedi2014modelbased} for Mckibben muscle actuated continuum robots, and a mechanics model was combined with passive cable displacement in \cite{rone2013passive} for shape sensing. Finally, string potentiometers were explored in \cite{trivedi2014modelbased,xu2017insertable,frazelle2018master} for shape sensing of continuum robots under planar or constant curvature shapes.
\par Despite previous works, there are two key technological and theoretical gaps that motivate this work. First, there is a need for a low cost solution to the problem of shape sensing. While use of FBG sensor arrays is possible, it is not an economical solution and seems to be at odds with the low cost of soft and continuum robots. Therefore, this paper is focused on presenting a modeling and design optimization framework for low-cost solutions to shape sensing of continuum robots using intrinsic sensing.  This solution assumes that the continuum/soft robot is equipped with actively actuated tendons (henceforth referred to simply as \emph{tendons}) and a set of passive measurement strings (henceforth referred to simply as \emph{strings}). These strings can be in the form of string potentiometers (akin to spring-loaded lanyards) or string encoders. The combined set of length measurements of these tendons and strings is used to estimate the shape of these robots. The second gap stems for the lack of a theoretical study focused on the effect of string routing on the shape sensing problem and design guidelines for string routing optimization to minimize the sensitivity of the shape sensing problem to measurement noise. To address this gap, we present a Lie group mathematical formulation that enables us to answer these core questions:
\begin{packed_itemize}
\item[\ding{227}] What is the impact of string routing on the kinematic sensitivity of the shape sensing to sensory noise?
\item [\ding{227}] How many sensors are needed per a continuum segment? and, if one has a fixed number of sensors, where should they be placed to minimize the shape error or to minimize the end effector error?
\end{packed_itemize}
\par While addressing these questions, we present a Lie group formulation for the string routing. We use a modal representation for the curvature distribution along the robot length and derive the general string routing kinematics relating variations in bending shape to string length. While one may obtain the string routing kinematics as in \cite{wang2019calibration} for calibration of continuum robots, we believe that the use of the Lie group formulation presented here leads to an elegant formulation of the body Jacobian. Finally, we adapt the noise amplification index presented by \cite{nahvi1996noise} within the context of robot calibration to guide the design of string routing.
\par Although accurate mechanics models have been presented in prior work on continuum robots, in this work we are interested in a formulation that could be used for online control, so we intentionally pursue a kinematics-based method, using lower-order polynomials that lead to either closed-form expressions or relatively cheap to compute optimization problems. Our kinematic deflection sensing approach does not require knowledge of applied loads from actuation, string encoders, or the environment, and it provides insight at the initial design stage for choosing string routings.
\par While our sensing approach does not rely on a mechanics model, our use of modal representation of curvature is related to continuum mechanics models for flexible \cite{odhner2012smooth} and continuum robots \cite{renda2020geometric,boyer2020dynamics}. Lie-group based kinematic integration formulations using the Magnus expansion have been used in other works for solving more general Cosserat rod mechanics models \cite{renda2020geometric,orekhov2020magnus} and a Lie-group finite-element formulation of beam dynamics was presented in \cite{JodickeMuller2014Lie_beam_dynamics}. Modal representations have also been used in prior works to parameterize the shape of hyper-redundant robots \cite{chirikjian1994modal} and continuum/soft robots \cite{sadati2017control}, with some works using modal functions to parameterize curvature/strain \cite{wang2014investigation,boyer2020dynamics} as we do here.
\par This paper is focused on shape sensing of larger scale hyper-redundant and continuum/soft robots, e.g. \cite{hannan2003elephant,mcmahan2006field,felt2019inductance}, but we believe the approach presented herein could also be applied to smaller continuum robots used in minimally-invasive surgery, e.g. \cite{ding2012design,sarli2019turbot}. This could potentially be achieved by mounting the string encoder housings remotely away from the robot body, as is already done with actuation units in surgical continuum robots.
\par The contributions of this paper lies in addressing the two key gaps identified above for single-segment continuum robots. In doing so, we present: 1) a Lie group kinematics formulation for shape sensing with general string routing and modal shape deflections, 2) an approach for optimizing string routing to improve the shape sensing kinematic conditioning with recommendations for designing string routings, and 3) an experimental validation and a simulation study of the proposed approaches on a torsionally stiff continuum robot and a torsionally compliant soft robot, respectively.
\section{Lie Group Kinematic Formulation}
This section presents the kinematic formulation and modal representation for parameterizing the variable curvature backbone shape and deriving the resulting pose of local frames along the continuum segment. This parametrization will be used for describing general string routing and the associated kinematic Jacobians for shape sensing.
\subsection{Central Backbone Kinematics}
\par
\begin{figure}[htbp]
  \centering
  \includegraphics[width=0.9\columnwidth]{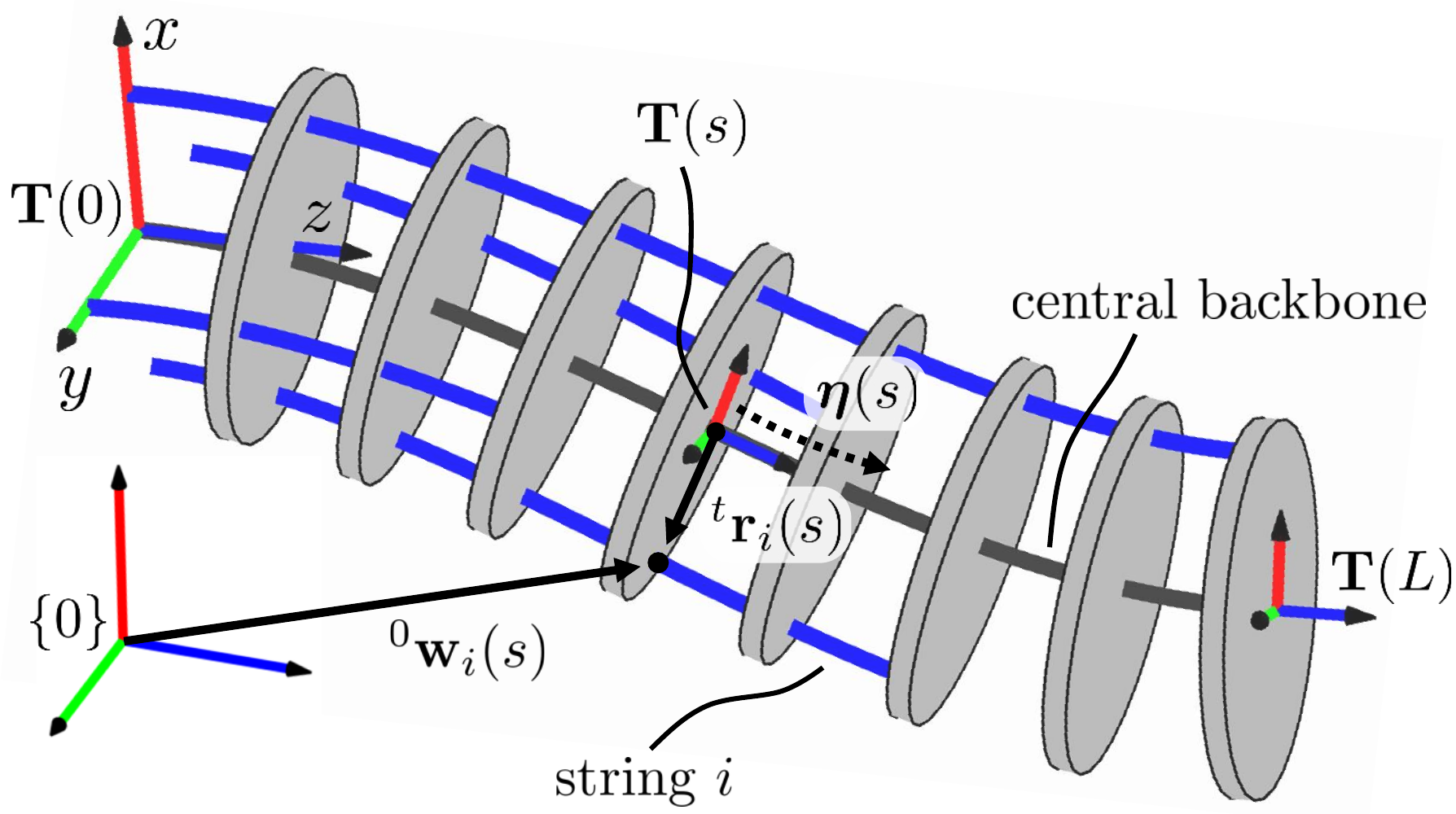}
  \caption{Variables used in our kinematic model to describe variable curvature deflections and general string routing.}
  \label{fig:kin_diagram}
\end{figure}

 Referring to Fig.~\ref{fig:kin_diagram}, we assume a robot with a total arc length $L$ and define the arc length coordinate $s$ as shown. We also assume that the robot's central backbone has a high slenderness ratio consistent with the assumption of negligible shear strains. With this assumption, the shape of the backbone can be described by its curvature distribution along its arc length $s$ in three directions, \mbox{$\mb{u}(s) = [u_x,u_y,u_z]\T \in \realfield{3}$}. For a given location, $s$, a local frame $\mb{T}(s)$ is assigned with its z-axis tangent and pointing in the direction of arc length growth and its two other axes in the backbone's local cross section:
\begin{equation} \label{eq:bb_frame_def}
\mb{T}(s) = \begin{bmatrix}
{}^0\mb{R}_t(s) & {}^0\mb{p}(s) \\
0 & 1
\end{bmatrix} \in SE(3), \quad s \in [0,L]
\end{equation}
where ${}^0\mb{R}_t(s)$ and ${}^0\mb{p}(s)$ are expressed in the world frame $\{0\}$\footnote{The notation ${}^a\mb{y}$ designates vector $\mb{y}$ described in frame \{A\} and $\Rot{a}{b}$ is the orientation of frame \{B\} with respect to frame \{A\} }. As the backbone changes its local curvature, this local frame undergoes a twist $\bs{\eta}(s)$ defined with the angular velocity preceding the linear velocity. The set of local frames associated with local curvatures $\mb{u}(s)$  has a corresponding twist distribution \mbox{$\bs{\eta}(s)$}. Describing the twist $\bs{\eta}(s)$ in its local frame $\mb{T}(s)$, we can write \mbox{$\bs{\eta}(s) = [\mb{u}(s)\T,\mb{e}_3\T]\T \in \realfield{6}$}  where $\mb{e}_3 = [0,0,1]\T$ denotes the local tangent unit vector.
\par As the frame $\mb{T}(s)$ undergoes the body twist $\bs{\eta}$, it satisfies the following differential equation \cite{lynch2017modern}:
\begin{equation} \label{eq:bb_frame_deriv}
\begin{gathered}
 \mb{T}'(s) = \mb{T}(s)\widehat{\bs{\eta}}(s), \quad \widehat{\bs{\eta}}(s) = \begin{bmatrix}
 \widehat{\mb{u}}(s) & \mb{e}_3 \\
 0 & 0
 \end{bmatrix} \in se(3) \\
  \widehat{\mb{u}}(s) =  \begin{bmatrix}
0 & -u_z(s) & u_y(s) \\
u_z(s) & 0 & -u_x(s) \\
-u_y(s) & u_x(s) & 0
 \end{bmatrix} \in so(3)
\end{gathered}
\end{equation}
where $\left( \cdot \right)'$ denotes the derivative with respect to $s$ and the hat operator $\left(\; \widehat{\cdot} \; \right)$ forms the standard matrix representations of $so(3)$ and $se(3)$ from their vector forms $\mb{u}$ and $\bs{\eta}$, respectively. We also define the adjoint representation of $se(3)$, which will be used below for computing Jacobians:
\begin{equation}
\text{ad}\left(\bs{\eta}(s)\right) = \begin{bmatrix}
\widehat{\mb{u}}(s) & \mb{0} \\
\widehat{\mb{e}}_3 & \widehat{\mb{u}}(s)
\end{bmatrix}
\end{equation}
\par In the following analysis, we choose to represent the curvature distribution $\mb{u}(s)$ as a weighted sum of polynomial functions (similar to \cite{renda2020geometric}). We denote the polynomial functions as $\bs{\phi}_x(s)$, $\bs{\phi}_y(s)$, and $\bs{\phi}_z(s)$ and the weights as $\mb{c}_x$, $\mb{c}_y$, and $\mb{c}_z$, for the $x$, $y$, and $z$ directions, respectively. The curvature distribution then takes the following form:
\begin{equation} \label{eq:curvature_basis}
\begin{aligned}
\mb{u}(s) = \begin{bmatrix} \bs{\phi}_x\T\mb{c}_x \\ \bs{\phi}_y\T\mb{c}_y \\ \bs{\phi}_z\T\mb{c}_z \end{bmatrix} &=
\begin{bmatrix}
\bs{\phi}_x\T & 0 & 0 \\
0 & \bs{\phi}_y\T & 0 \\
0 & 0 & \bs{\phi}_z\T
\end{bmatrix} \begin{bmatrix} \mb{c}_x \\ \mb{c}_y \\ \mb{c}_z \end{bmatrix} \\
&= \bs{\Phi}(s)\mb{c}, \quad \bs{\Phi}(s) \in \realfield{3\times{}m}, \quad \mb{c} \in \realfield{m}
\end{aligned}
\end{equation}
where the columns of $\bs{\Phi}(s)$ form a \emph{modal shape basis}, and $\mb{c}$ is a vector of constant \emph{modal coefficients}.
\par The modal description of the curvature distribution offers a description of a variety of variable curvature deflections using a finite set of modal coefficients.  This also provides simplified kinematic expressions for computing the workspace, solving for the shape using the string encoder measurements, and designing the string encoder routing to improve the numerical conditioning of the shape sensing problem.
\par For a given configuration $\mb{c}$, the frames $\mb{T}(s)$ are found by integrating (\ref{eq:bb_frame_deriv}). A number of Lie group integration methods could be used for this, as reviewed in \cite{iserles2000liegroup}, but here we use an approach based on the Magnus expansion because both fourth and sixth order expansions can be computed efficiently and because of the Magnus expansion's large convergence bound \cite{iserles2000liegroup}. After integration with a Magnus expansion method (or another Lie group integration method), the spatial curve is given as a product of matrix exponentials \cite{orekhov2020magnus}:
\begin{equation} \label{eq:prod_of_exp}
\mb{T}(s) = \mb{T}(0)\prod_{i=0}^{k} e^{\mb{\Psi}_i}, \quad \mb{\Psi}_i \in se(3)
\end{equation}
\par Details on computing $\mb{\Psi}_i$ can be found in \cite{iserles2000liegroup,orekhov2020magnus}. We will show below that because we use a modal shape basis, Lie group integration is not needed to solve the shape sensing problem (i.e. determining the modal coefficients), however, once the configuration $\mb{c}$ is determined, (\ref{eq:bb_frame_deriv}) must be integrated to compute the robot's forward kinematics and Jacobian, as shown in Section \ref{sec:body_jacobian}. In this paper, we compute \eqref{eq:prod_of_exp} with small step sizes (100 points along the backbone) at a rate of $\sim 45$ Hz in an unoptimized MATLAB implementation. Larger steps can be used in practice for faster computation times\cite{orekhov2020magnus}, but we use a fine discretization here to avoid introducing additional integration error into this study.
\subsection{Modal Shape Basis with Chebyshev Polynomials}
\par
\begin{figure}[tbp]
  \centering
  \includegraphics[width=0.99\columnwidth]{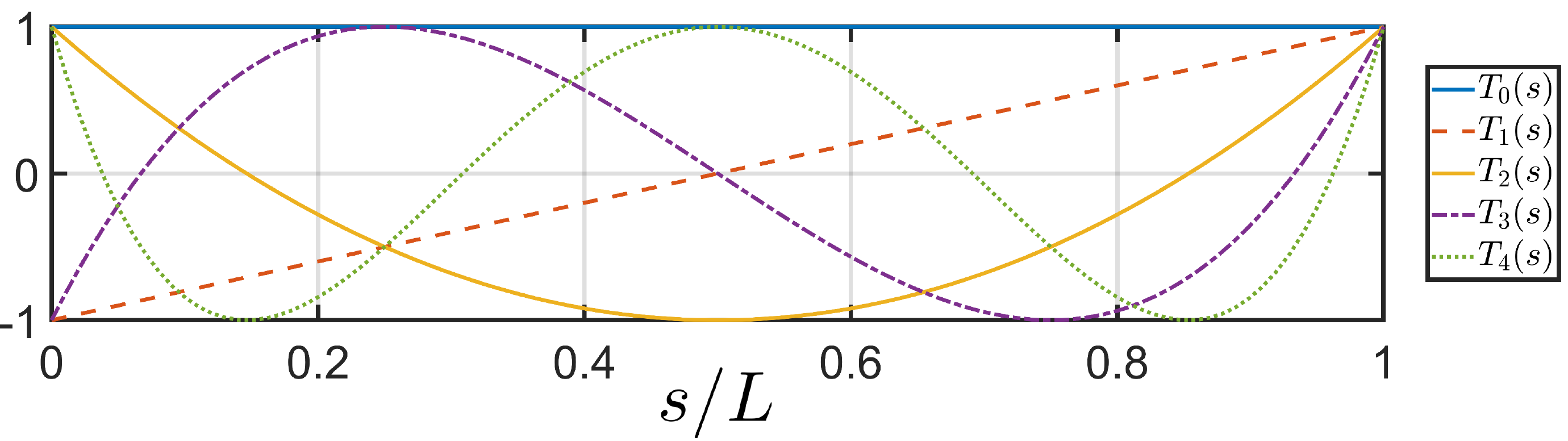}
  \caption{The first five shifted Chebyshev polynomials.}
  \label{fig:cheby_polynomials}
\end{figure}

 Although any set of modal shape functions could be used to form $\bs{\Phi}$, (e.g. Euler curves \cite{gonthina2019euler,rao2021euler}, monomials \cite{wang2019calibration}, Legendre polynomials \cite{odhner2012smooth}, or trigonometric functions), in this paper we use Chebyshev polynomials of the first kind since each polynomial is bounded by $\pm{}1$, which provides improved scaling of the modal coefficients and simplifies computation of the admissible workspace, as described in Section \ref{sec:routing_design}. The Chebyshev polynomials can be expressed recursively as \cite{mason2002chebyshev}:
\begin{equation}\label{eq:cheby_recursive}
\begin{gathered}
T_0 = 1, \quad T_1(x) = x\\
T_n(x) = 2xT_{n-1}(x)-T_{n-2}(x), \quad n = 2,3,...
\end{gathered}
\end{equation}
where $T_n(x), \; x \in [-1,1]$ is the $n^{th}$ degree Chebyshev polynomial. Since the arc length coordinate is given by $s \in [0,L]$ and $T_n(x)$ is defined for $ x \in [-1,1]$, the following coordinate transformation is used:
\begin{equation} \label{eq:cheby_transformation}
x(s) = \frac{2s - L}{L}
\end{equation}
We then evaluate $T_n(x(s))$ via \eqref{eq:cheby_recursive}. In the remainder of this paper, we refer to the polynomials as simply $T_n(s)$ with the transformation \eqref{eq:cheby_transformation} implied. The first five Chebyshev polynomials, shifted to $s \in [0,L]$, are shown in Fig. \ref{fig:cheby_polynomials}.
\par As an example, to represent a $y$ direction curvature with a second-order Chebyshev series, the modal shape basis and modal coefficients would be given by:
\begin{equation} \label{eq:planar_y_2nd_order_basis}
\begin{gathered}
u_y(s) = \bs{\phi}_y(s)\mb{c}_y \\
\bs{\phi}_y\T(s) = \begin{bmatrix}
T_0 & T_1(s) & T_2(s)
\end{bmatrix}\T, \quad \mb{c}_y \in \realfield{3}
\end{gathered}
\end{equation}
where the first three Chebyshev polynomials, shifted to $s \in [0,L]$ are given by:
\begin{equation} \label{eq:cheby_first_three}
T_0 = 1, \quad T_1(s) = \frac{2s - L}{L}, \quad T_2(s) = \frac{8s^2}{L^2} - \frac{8s}{L} + 1\\
\end{equation}
\begin{figure}[tbp]
  \centering
  \includegraphics[width=0.99\columnwidth]{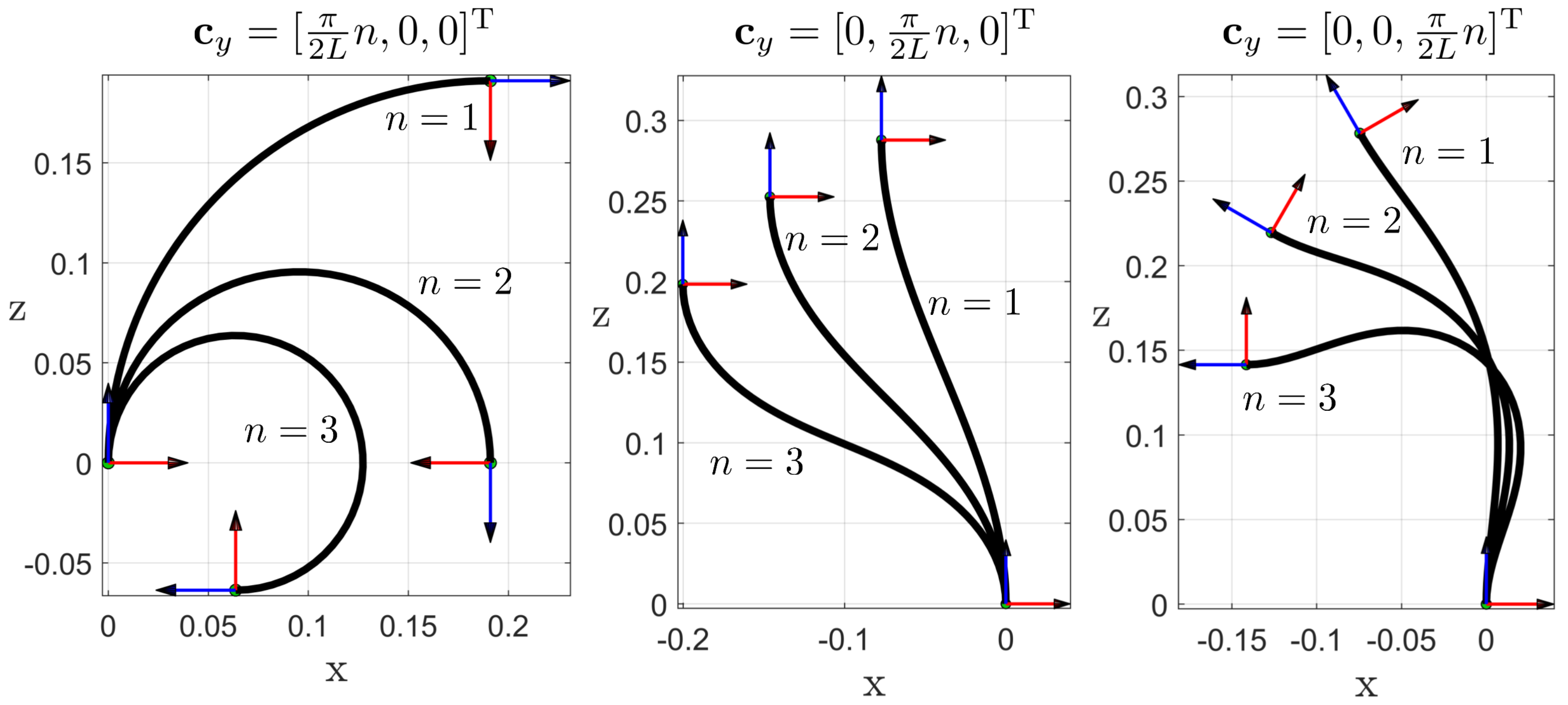}
  \caption{Deflections generated on a rod with $L = 300$ mm and second order Chebshev series modal shape basis on the $y$ direction curvature, as given by (\ref{eq:planar_y_2nd_order_basis}). Deflections were generated by taking $\frac{\pi}{2L}$ steps in each of the three modal coefficient directions.}
  \label{fig:cheby_shapes}
\end{figure}
\par Figure \ref{fig:cheby_shapes} illustrates shapes generated by these first three Chebyshev polynomials for a planar continuum segment. Of note is that shapes in the $T_0$ and $T_1$ directions of the modal shape basis correspond to experimentally observed deflection shapes in tendon-actuated and multi-backbone robots. Shapes along the $T_0$ direction correspond to constant curvature deflections exhibited in continuum segments with actuation wires equidistantly distributed about the central backbone. Shapes along the $T_1$ direction correspond to constant-orientation deflections due to external forces applied to the end disk of a tendon-actuated continuum segment, as shown in Fig. \ref{fig:deflection_examples}.
\par The shape sensing methods presented below would apply to other choices for $\bs{\Phi}$ as well. For example, a modal shape basis with coupling between the modal coefficients could be used, e.g. \cite{renda2020geometric}. If we choose $\mb{\Phi}$ as the identity matrix, $\mb{c}$ corresponds directly to constant curvatures in the $x$, $y$, and $z$ directions (see Fig. \ref{fig:cheby_shapes}), so the results here also apply to robots modeled with the commonly used constant-curvature assumption \cite{webster2010cc_review}.
\subsection{General String Routing Kinematics}
In \cite{rucker2011statics}, a Cosserat-rod based mechanics model was presented for tendon-actuated robots with general tendon routing. In the kinematics model presented here, we use a similar method for describing the string routing. Assuming $p$ strings, the string path, as shown in Fig. \ref{fig:kin_diagram}, is expressed in the moving frame $\mb{T}(s)$ and given by:
\begin{equation} \label{eq:string_route_def}
{}^t\mb{r}_i(s) = [ r_{x_i}(s), r_{y_i}(s), 0] \T, \quad i = 1,2,\dots,p
\end{equation}
Our kinematic formulation permits any differentiable function for $\mb{r}(s)$, but in Sections \ref{sec:experiments_mfg} and \ref{sec:experiments_soft} we consider constant pitch-radius paths and helical paths. The position of a point along the string/wire rope path in the world frame is given as the vectorial sum of the point $\mb{p}(s)$ along the central backbone and the radial vector $\mb{r}(s)$, defined in the moving frame:
\begin{equation} \label{eq:string_path}
{}^0\mb{w}_i(s) = {}^0\mb{p}(s) + \Rot{0}{t}(s)\,{}^t\mb{r}_i(s)
\end{equation}
Noting that vector norms are invariant under rotations, the length of the $i^{th}$ string is given by:
\begin{equation} \label{eq:string_length}
\ell_i = \int_{0}^{s_{a_i}} \|{}^t\mb{w}'_i(s) \| \, ds, \quad i = 1,2,\dots,p \\
\end{equation}
where $s_{a_i}$ designates arc length along the central backbone at which the string is anchored to the spacer disk/end disk. Taking the derivative of \eqref{eq:string_path} with respect to $s$, substituting \eqref{eq:curvature_basis}, and then using $ {}^t\mb{w}'_i(s) = \Rot{0}{t}\T {}^0\mb{w}'_i(s)$ results in:
\begin{equation} \label{eq:string_path_deriv}
{}^t\mb{w}'_i(s)= \mb{e}_3 - {}^t\widehat{\mb{r}}_i(s)\mb{\Phi}(s)\mb{c}  + {}^t\mb{r}_i'(s), \quad i=1,2,\ldots p
\end{equation}
Deriving the above result also requires \eqref{eq:bb_frame_deriv}, which states that \mbox{${}^0\mb{p}'(s) = \Rot{0}{t}(s)\mb{e}_3$} and ${}^0\mb{R}'_t(s) = \Rot{0}{t}(s)\widehat{\mb{u}}(s)$.
\par Recalling that ${}^t\mb{r}_i(s)$ is in a moving frame having a body angular velocity $\widehat{\mb{u}}(s)$, we can visualize (\ref{eq:string_path_deriv}) as the velocity of a point traversing the string path as the arc length $s$ is increased at a unit speed. This point speed is given as the sum of the induced velocity due to the rotation of the moving frame and the velocity of that point relative to the moving frame due to traversal along the central backbone and the rate of change of the string's radial placement ${}^t\mb{r}_i'(s)$. Since (\ref{eq:string_path_deriv}) is a function only of the modal basis, the string routing function, and the modal coefficients, we can numerically integrate (\ref{eq:string_length}) with any quadrature rule (e.g. the trapezoid rule) and avoid the cost of integrating the Lie group differential equation in (\ref{eq:bb_frame_deriv}).
\par The string routings ${}^t\mb{r}_i, \; i=1\ldots p$ can be chosen by the designer, but must each satisfy a geometric constraint that the string path in world frame must not have a cusp at any configuration of the continuum segment.  To ensure physically realizable string paths, the local tangent to the actuation string must point in the same direction as the local tangent of the central backbone. It is possible for the angular rate of changes $u_x$ and $u_y$ to be large enough to cause ${}^t\mb{w}'_i$ to point in the opposite direction of $\mb{e}_3$, which causes the string path to change directions. To avoid this scenario, we  require that ${}^t\mb{w}'_i$ always point in the same direction as $\mb{e}_3$. Using \eqref{eq:string_path_deriv}, we obtain the following set of $p$ constraint equations for each string:
\begin{equation} \label{eq:realizable_path_constraint}
\left({}^t\mb{w}'_i\right)\T\mb{e}_3 = r_{y_i}(s)u_x(s) - r_{x_i}(s)u_y(s) + 1 > 0
\end{equation}
This can be rewritten as constraints on $u_x$ and $u_y$:
\begin{equation} \label{eq:realizable_curvature_bounds}
\begin{aligned}
u_x(s) &\leq \frac{r_{x_i}(s)u_y(s) + 1}{r_{y_i}(s)}, \quad u_y(s) \leq \frac{r_{y_i}(s)u_x(s) + 1}{r_{x_i}(s)} \\
\end{aligned}
\end{equation}
In the coming sections, the string routings will be incorporated into a model that captures their effect on increasing the robustness to noise in string length measurements to changes in the estimated shape of the continuum segment.
\subsection{Solving for the Modal Coefficients} \label{sec:solving_shape}
To solve for the shape of the robot, we concatenate \eqref{eq:string_length} for each string together with the string length measurements $\bs{\ell}^*$:
\begin{equation} \label{eq:shape_sense_problem}
\bs{\ell}(\mb{c}) - \bs{\ell}^* = 0, \quad \bs{\ell} \in \realfield{p}, \quad \bs{\ell}^* \in \realfield{p}
\end{equation}
and solve this system of equations for the modal coefficients $\mb{c}$. We then integrate $\mb{T}'(s)$ once to find the backbone pose at any desired arc length. In some cases, for particular choices of ${}^t\mb{r}_i$ and $\mb{\Phi}$, unique closed-form solutions to (\ref{eq:shape_sense_problem}) can be found by explicit integration of (\ref{eq:string_length}). This occurs below for the planar case and for the case of robots with high torsional stiffness.
\par In general however, the system of equations in (\ref{eq:shape_sense_problem}) can be nonlinear in $\mb{c}$ and may have multiple solutions, so it must be solved by an iterative numerical method, e.g. Gauss-Newton. A necessary condition for using the Gauss-Newton algorithm is $p \geq m$ where $p$ is the number of strings and $m$ is the number of columns in $\bs{\Phi}$. The Jacobian needed for each iteration of the Gauss-Newton method is provided in the next section.
\subsection{Configuration Space Jacobian} \label{sec:config_jacobians}
\par Since the modal vector $\mb{c}$ uniquely defines the shape of the continuum segment for a given modal basis $\bs{\Phi}$, we use $\mb{c}$ as the configuration space variable. We also define the \emph{configuration-space Jacobian} as the Jacobian relating small changes in the string lengths to small changes in the modal coefficients:
\begin{equation} \label{eq:config_jacobian}
d\bs{\ell} = \mb{J}_{\ell{}c}d\mb{c}, \quad \mb{J}_{\ell{}c} \in \realfield{p \times m}
\end{equation}
Following \cite{boyer2020dynamics}, the $i^{th}$ row of $\mb{J}_{\ell{}c}$ can be derived as:
\begin{equation} \label{eq:config_jacobian_row}
\frac{d\ell_i}{d\mb{c}} = \int_0^{s_{a_i}}  \left(  {{}^t\mb{r}_i} \times \frac{ \left({}^t\mb{w}'_i \right)}{\|  {}^t\mb{w}'_i \| } \right)  \T  \bs{\Phi} \;  ds \\
\end{equation}
where we have dropped the dependence on $s$ in the integrand  terms for brevity and ${}^t\mb{w}'_i $ was given in \eqref{eq:string_path_deriv}. As with the integral in (\ref{eq:string_length}), (\ref{eq:config_jacobian_row}) can be computed with any quadrature rule.
\subsection{Body Jacobian} \label{sec:body_jacobian}
\par Next, we define the \emph{body Jacobian} as the Jacobian relating the twists of the moving frame $\mb{T}(s)$  expressed in  $\mb{T}(s)$ (also called body twists) to small changes in the modal coefficients:
\begin{equation} \label{eq:body_jacobian}
\bs{\xi}(s) = \mb{J}_{\xi{}c}(s)d\mb{c}, \quad \bs{\xi} \in se(3)
\end{equation}
where $\bs{\xi}(s) = [\bs{\omega}(s)\T ,\mb{v}(s)\T]\T$ is the instantaneous twist of $\mb{T}(s)$ produced by $d\mb{c}$, expressed in the body frame $\mb{T}(s)$ and with angular velocity followed by linear velocity. The $i^{th}$ column of $\mb{J}_{\xi{}c}(s)$, denoted by $\mb{J}^{[i]}_{\xi{}c}(s)$ is the twist produced by a small change in the $i^{th}$ element of $\mb{c}$, denoted by $c_i$:
\begin{equation} \label{eq:task_jacobian_column}
\mb{J}^{[i]}_{\xi{}c}(s) = \left(\mb{T}^{-1}(s) \frac{\partial}{\partial{}c_i}\left(\mb{T}(s)\right)\right)^\vee \in se(3)
\end{equation}
where $(\cdot)^\vee$ denotes the inverse operation of $\left(\; \widehat{\cdot} \; \right)$.
\par Referring back to \eqref{eq:prod_of_exp}, we require the terms $\frac{\partial}{\partial{}c_i}\left(e^{\bs{\Psi}_j}\right)$ to be able to compute $\frac{\partial}{\partial{}c_i}\left(\mb{T}(s)\right)$. These derivatives are given by:
\begin{equation}
\frac{\partial}{\partial{}c_i} e^{\mb{\Psi}_j} = e^{\mb{\Psi}_j} \text{dexp}\left( \frac{\partial{}\mb{\Psi}_j}{\partial{}c_i}\right)
\end{equation}
where the $\textrm{dexp}$ operator is defined in \cite{rossmann2006lie} as:
\begin{equation}
\text{dexp}\left( \frac{\partial{}\mb{\Psi}_j}{\partial{}c_i}\right) = \left( \sum_{k=0}^\infty \frac{\left(-1\right)^k}{\left( k+1 \right) !} \text{ad}^k\left( \mb{\Psi}_j \right) \right) \left( \frac{\partial{}\mb{\Psi}_j}{\partial{}c_i}\right)
\end{equation}
\par Closed-form expressions for $\text{dexp}$ for the case of $se(3)$ are also available in \cite{selig2004geometric}. The value of $\frac{\partial{}\mb{\Psi}_j}{\partial{}c_i}$ can be derived symbolically and will vary depending on the order of the Magnus expansion used. Alternatively, it can be estimated via finite difference approximation.
\par The kinematic expressions defined above provide the equations needed for solving the shape sensing problem and computing the forward/inverse kinematics for a particular continuum robot design and string routing function definition. In Section \ref{sec:experiments_mfg} we provide experimental validations of this modeling and shape sensing approach.
\section{String Routing Optimization} \label{sec:routing_design}
\par A benefit of our proposed kinematic formulation is that in addition to capturing variable curvature deflections, it provides a designer flexibility in the design of the string routing by allowing non-straight string routings. In some cases, there may be practical mechanical integration considerations that would benefit from non-straight routing, e.g. routing strings around proprioceptive sensing electronics embedded in the continuum structure \cite{abah2022sensors} or routing strings in a tapered path for a segment with a decreasing diameter from base to tip \cite{renda2014dynamic}. In this section, we provide considerations for choosing the string routing paths to reduce the propagation of string measurement error to pose error while avoiding ill-conditioned Jacobians.
\par For a general purpose manipulator, the external loading magnitude and location may not be known \emph{a priori}. For this reason, here we propose a Jacobian-based method to optimize the string routing path ${}^t\mb{r}_i(s)$ without assuming any particular loading conditions. We do this by designing ${}^t\mb{r}_i(s)$ to improve the numerical conditioning of both the task space and configuration space Jacobians to reduce the upper bound on error propagation from the string length measurements to errors in the spatial curve $\mb{T}(s)$. We validate the approach in simulations and experiments in Sections \ref{sec:planar_example}, \ref{sec:experiments_mfg}, and \ref{sec:experiments_soft}.
\par We first define the noise amplification index, a design measure used in robot calibration \cite{nahvi1996noise}. For an expression $ \mb{A}\mb{x} = \mb{b}$ the noise amplification is given by:
\begin{equation}
\aleph \left( \mb{A} \right) = \frac{\sigma^2_{min}\left( \mb{A} \right)}{\sigma_{max}\left( \mb{A} \right)}
\end{equation}
where $\sigma_{min}\left( \mb{A} \right)$ and $\sigma_{max}\left( \mb{A}\right)$ are the minimum and maximum singular values of $\mb{A}$, respectively. As shown in \cite{nahvi1996noise}, the noise amplification index provides a bound on how errors in $\mb{b}$ propagate to errors in $\mb{x}$:
\begin{equation}\label{eq:noise_amp_bound}
\| \delta \mb{x} \| \leq \frac{1}{\aleph \left( \mb{A} \right)} \| \delta \mb{b} \|
\end{equation}
\par Within the context of shape sensing, \eqref{eq:config_jacobian} is the mapping relating noise in string length measurement $d\bs{\ell}$ to a change in modal coefficients $d\mb{c}$. Also, \eqref{eq:body_jacobian} provides the mapping relating twist and changes in $d\mb{c}$. We solve \eqref{eq:config_jacobian} for $d\mb{c}$ and substitute into \eqref{eq:body_jacobian}, then solve the equation for $d\bs{\ell}$ to produce:
\begin{equation} \label{eq:full_kin_map}
d\bs{\ell} = \mb{J}_{\ell\xi}\bs{\xi}(s), \quad \mb{J}_{\ell\xi} = \left( \mb{J}_{\xi c} \mb{J}^+_{\ell c} \right)^+, \quad
\end{equation}
where $\left(^+\right)$ denotes the pseudoinverse. Using \eqref{eq:noise_amp_bound}, \eqref{eq:full_kin_map} has the following noise amplification bound:
\begin{equation}\label{eq:twist_sensitivity}
||\delta \bs{\xi}(s)||\leq \frac{1}{\aleph\left( \mb{J}_{\ell\xi} \right) }||\delta d\bs{\ell}||
\end{equation}
\par Since the pose $\mb{T}(s)$ is an integral of the twist, minimizing the effect of string encoder measurement noise on the estimates of the segment shape requires minimizing $||\delta \bs{\xi}(s)||$, which in turn requires maximizing $\aleph(\mb{J}_{\ell\xi})$. Since $\aleph(\mb{J}_{\ell\xi})$ contains linear/angular velocity units, we multiply the first three rows corresponding to angular velocity by a characteristic length $c_\ell$ \cite{cardou2010kinematic}. For our experimental and simulation results, we chose $c_\ell$ to be the segment's kinematic radius so that angular velocities are scaled to represent linear velocities at the edge of the disk.
\par We must also consider the numerical conditioning of the configuration space Jacobian $\mb{J}_{\ell c}$, since it is used when iteratively solving (\ref{eq:shape_sense_problem}). Since maximizing $\aleph(\mb{J}_{\ell\xi})$ does not guarantee a well-conditioned $\mb{J}_{\ell c}$, we seek to also prevent $\aleph(\mb{J}_{\ell c}) = 0$, which would indicate a singular $\mb{J}_{\ell c}$. We define this design problem as a constrained optimization problem:
\begin{equation} \label{eq:design_problem}
\max_{\mb{k}} \; \aleph_g\left(\mb{J}_{\ell\xi} \right)\quad
\textrm{s.t.} \; \;  \aleph\left( \mb{J}_{\ell c} \right) \geq \epsilon
\end{equation}
where $\aleph_g$ is the global noise amplification index defined below, $\mb{k}$ are a set of string path design parameters, and $\epsilon$ is a lower-bound on the noise amplification index. We will provide two examples of defining $\mb{k}$ to ensure the string paths are physically realizable in the simulation and experimental studies below. For these two examples, $\mb{k}$ are restricted to a set of integers such that \eqref{eq:design_problem} can be solved by a brute-force search.
\par We define $\aleph_g(\mb{A})$ as the global noise amplification index, which is the average of $\aleph(\mb{A})$ over an admissible configuration workspace $\mathcal{C}_a$ denoting all admissible configurations $\mb{c}$. This global performance measure can be numerically approximated by sampling $\mathcal{C}_a$ along $j$ sample points $\mb{c}_1\ldots\mb{c}_j$ and computing $\aleph(\mb{A})$ at each sampled configuration:
\begin{equation}
\aleph_g(\mb{A}) \triangleq \frac{\int_{\mathcal{C}_a}^{}\aleph(\mb{A}) \, d\mathcal{C}_a}{\int_{\mathcal{C}_a}^{}d\mathcal{C}_a} \approx \frac{1}{j}\sum_{i=1}^{j}\aleph(\mb{A}_i)
\end{equation}
\par We define the admissible workspace $\mathcal{C}_a$ as the set of all shapes that a segment can achieve:
\begin{equation}
\mathcal{C}_a = \{ \mb{c}\; | \; \mb{f}(\mb{c}) \leq \mb{0} \}
\end{equation}
where $\mb{f}(\mb{c})$ is a vector of constraints on the robot's configuration. To compute the admissible workspace, we take samples in the configuration space $\mb{c}$ and discard samples that violate the constraints $\mb{f}(\mb{c})$. The constraints $\mb{f}(\mb{c})$ need to be defined on a case-by-case basis, but in this paper we define three constraints that are relevant to the robots considered in our simulation studies and experimental results.
\par The first set of constraints we consider are maximal strain limits on the continuum structure $\bs{\epsilon}_{max} = [\epsilon_{max,x},\epsilon_{max,y},\epsilon_{max,z}]\T$. Assuming a backbone diameter $d_b$ and a beam model following linear elasticity, the local curvature limits are given by:
\begin{equation}\label{eq:strain_constraint}
\text{max}_s\left( u_j(\mb{c}, s) \right) - \frac{\epsilon_{max,j}}{(d_b/2)} \leq 0, \quad j\in\{x, y, z\}
\end{equation}
where $\text{max}_s$ denotes the maximum over $s \in [0,L]$ for a configuration $\mb{c}$.
\par The second set of constraints we consider are the maximum curvatures to prevent the intermediate disks from colliding.
\begin{figure}[htbp] \centering
  \includegraphics[width=0.8\columnwidth]{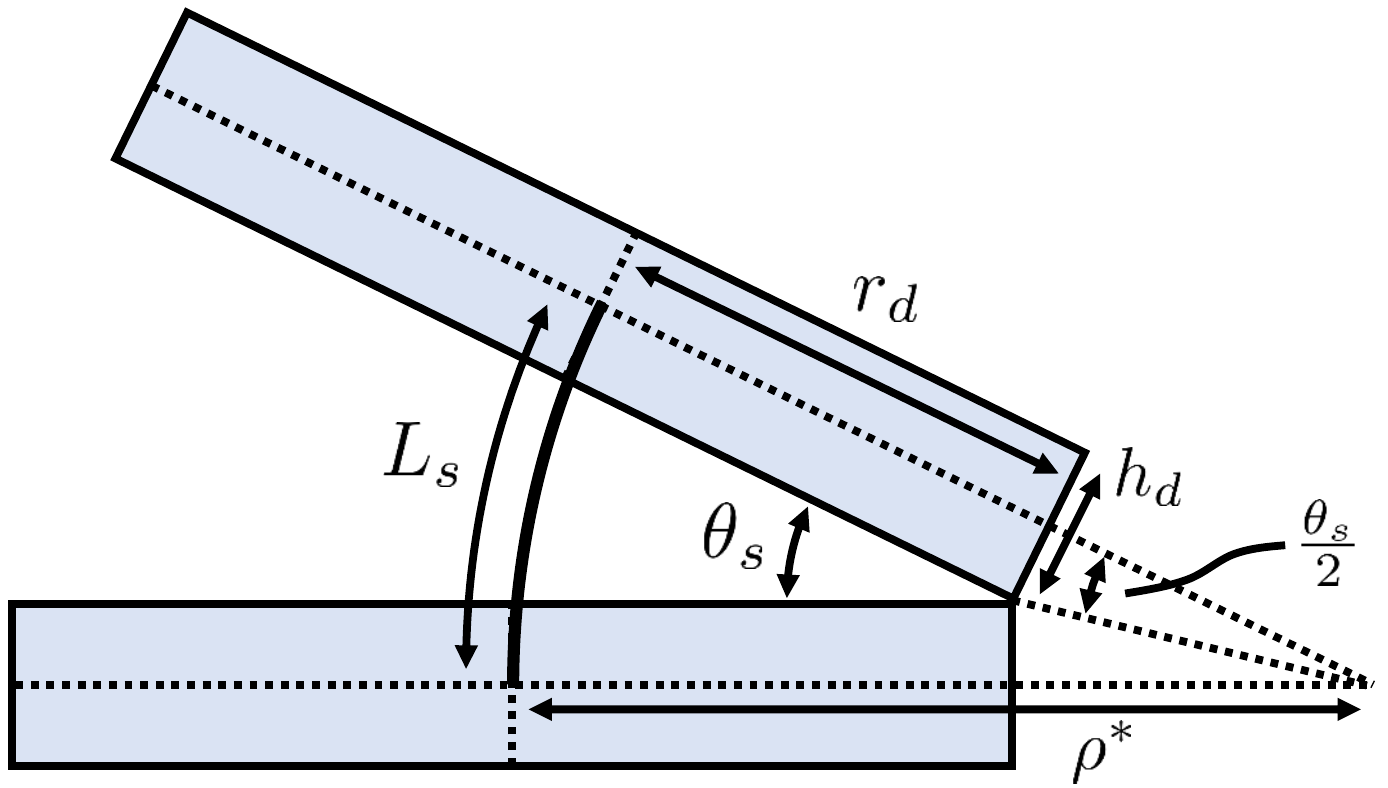}
  \caption{Kinematic variables for a subsegment in a constant curvature configuration with disk collision.}
  \label{fig:disk_collision}
\end{figure}
Considering a subsegment of the robot between two intermediate disks with length $L_s$, and assuming the subsegment is under constant curvature as shown in Fig. \ref{fig:disk_collision}, we have:
\begin{equation} \label{eq:subseg_triangle}
\text{tan}\left(\frac{\theta_s}{2}\right) = \frac{h_d}{2\left(\rho^* - r_d\right)}
\end{equation}
where $\theta_s$ is the angle between the intermediate disks, $h_d$ is the height of each intermediate disk, $r_d$ is the radius of each disk, and $\rho^*$ is the radius of curvature when the disks collide. Substituting $\theta_s = \frac{L_s}{\rho^*}$ into \eqref{eq:subseg_triangle}, we have the following
\begin{equation} \label{eq:disk_collision}
2\left(\rho^* - r_d\right)\text{tan}\left( \frac{L_s}{2\rho^*}\right) = h_d
\end{equation}
which we solve numerically for $\rho^*$. To prevent disk collision, we then require that the curvatures in the $x$ and $y$ direction are low enough to avoid this collision condition:
\begin{equation}\label{eq:disk_collision_constraint}
\text{max}_s\left( \| \mb{u}_{x/y}(s)\| \right) \leq \frac{1}{\rho^*}
\end{equation}
where $\mb{u}_{x/y}(s) = [u_x(s),u_y(s)]\T$ and $\text{max}_s$ again denotes the maximum over $s \in [0,L]$.
\par The third set of constraints we include are given by \eqref{eq:realizable_curvature_bounds}, which ensure that the string paths are physically realizable. Using these three sets of constraints (as specified by \eqref{eq:strain_constraint}, \eqref{eq:realizable_curvature_bounds}, and \eqref{eq:disk_collision_constraint}), we determine the admissible workspace by searching for configurations that do not violate the constraints, and compute $\aleph_g$ for these admissible configurations. This allows us to explore how to design the string routing paths to maximize $\aleph_g$, which we do below for several simulation and experimental examples.
\section{Planar Case Study} \label{sec:planar_example}
\par In this section, we present a simulation case study for a segment subject only to planar deflections. We show that for planar deflections and string routings with constant pitch radius, the configuration space Jacobian $\mb{J}_{\ell c}$ is constant for any choice of the modal basis $\bs{\Phi}(s)$. We then explore choices of string anchor points and pitch radii that improve the noise amplification indices $\aleph_g \left( \mb{J}_{\ell\xi}\right)$ and  $\aleph \left( \mb{J}_{\ell{}c}\right)$.
\begin{figure}[htbp]
  \centering
  \includegraphics[width=0.99\columnwidth]{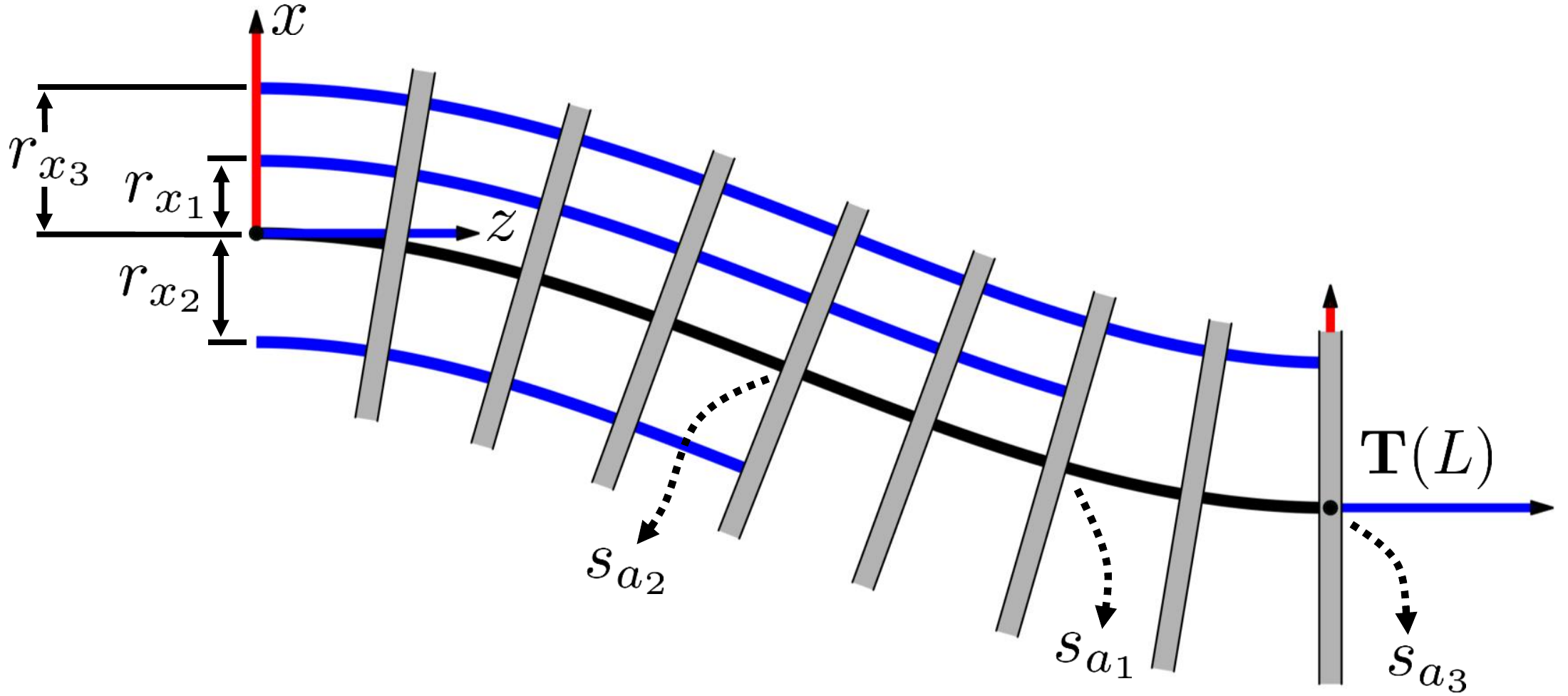}
  \caption{Variables of the kinematic model used in the planar case study.}
  \label{fig:planar_kinematics}
\end{figure}
\par Consider a planar continuum segment that is restricted to deflect in the $x-z$ plane, i.e. $u_x = u_z = 0$, as shown in Fig. \ref{fig:planar_kinematics}. We choose to represent the curvature distribution using a second-order Chebyshev series as given in \eqref{eq:planar_y_2nd_order_basis}, and we assume three strings are routed within the segment and anchored at arc lengths $s_{a_1}$, $s_{a_2}$, and $s_{a_3}$. We also assume the strings are routed in a path with a constant pitch radius, i.e. ${}^t\mb{r}_i(s) = [r_{x_i},0,0]\T$, where $r_{x_i} \in \realfield{}$ is a constant scalar. Noting that ${}^t\mb{r}_i'(s) = [0,0,0]\T$, \eqref{eq:string_path_deriv} simplifies to:
\begin{equation}
{}^t\mb{w}'_i(s) = \left[0,\, 0,\, \left(1- r_{x_i}\bs{\phi}_y\T(s)\mb{c}_y \right) \right]\T
\end{equation}
Applying the requirement from \eqref{eq:realizable_path_constraint} that $\left({}^t\mb{w}'_i\right)\T\mb{e}_3 > 0$, for this planar case, \eqref{eq:string_length} simplifies to:
\begin{equation} \label{eq:planar_single_string_length}
\ell_i = s_{a_i} - r_{x_i}\int_0^{s_{a_i}}\bs{\phi}\T_y(s) \mb{c}_y \, ds
\end{equation}
\par We will now consider how many strings are needed to accurately predict the tip pose $\mb{T}(L)$ of this planar segment. We assume here that the number of columns in the modal shape basis is equal to the number of strings, i.e. $p = m$ and $\mb{J}_{\ell c}$ is square. This means each additional sensing string enables an additional higher-order term to be added to the shape basis to further reduce the tip pose error. Prior works that used shape functions for modeling continuum robots have shown that low-order shape functions can be sufficient for capturing variable curvature deflections \cite{zhang2009JMED_cochlea,gonthina2019euler,renda2020geometric}. We will also demonstrate this here for this simulation study and experimentally in Sections \ref{sec:experiments_mfg} and \ref{sec:experiments_soft}.
\par
\begin{figure}[ht]
  \centering
 \includegraphics[width=0.95\columnwidth]{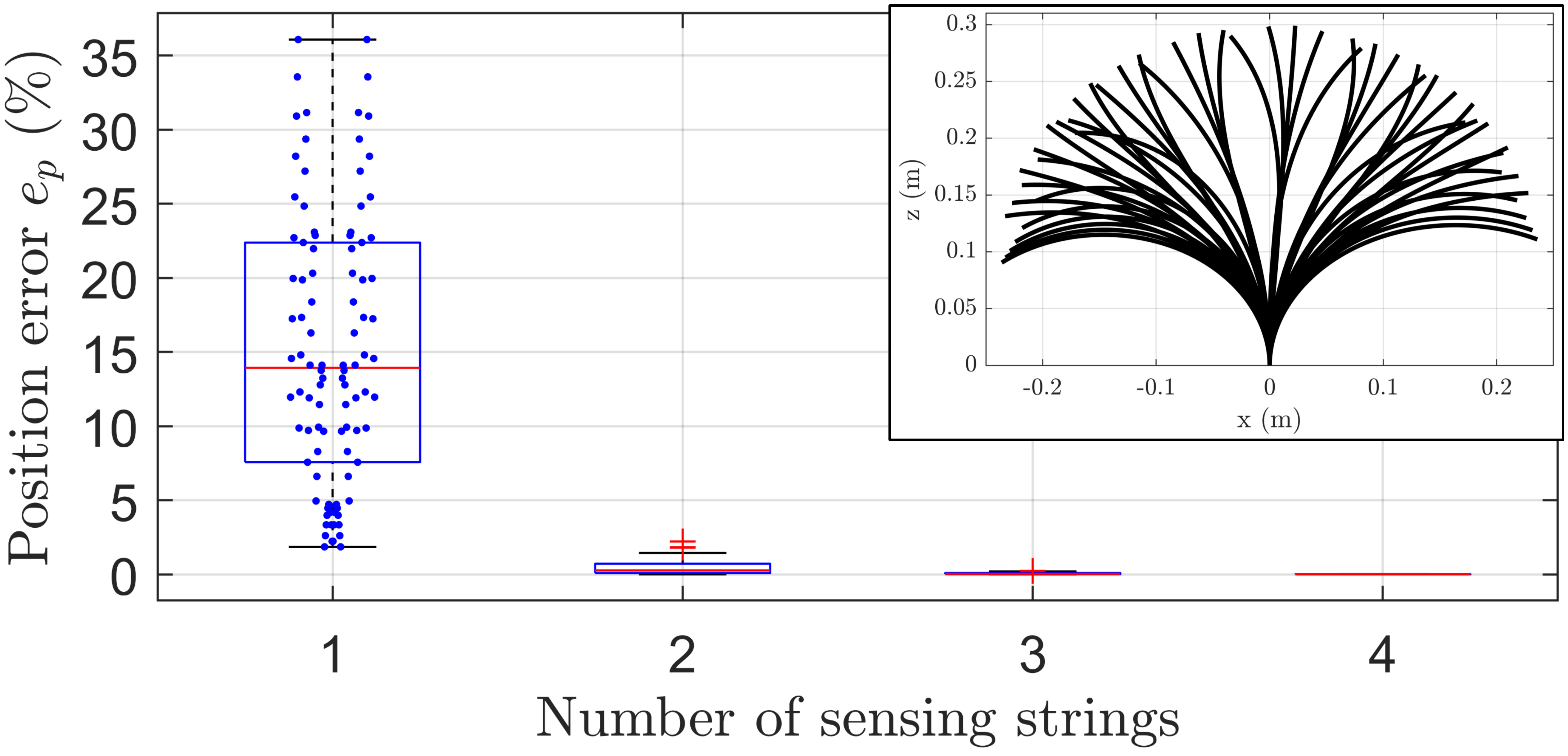}
  \caption{The tip position error between our kinematic model and a simulated Nitinol rod rapidly converges as the number of strings $p$ is increased. Individual data points are shown for $p=1$, and a subset of the 100 simulated variable curvature rod shapes is shown in the inset. }
  \label{fig:planar_convergence_boxplot}
\end{figure}

 We used a Cosserat rod mechanics model from \cite{orekhov2020magnus}, which neglected shear strains and extension, to simulate a Nitinol rod with a length $L = 300$ mm and a diameter of 4 mm (similar to the central backbone Nitinol rod used in the robot in Section \ref{sec:experiments_mfg}). To generate a variety of variable curvature rod shapes, we subjected the rod to planar forces in the world frame's $x$ direction and moments in the world frame's $y$ direction, with the world frame assigned as shown in Fig. \ref{fig:planar_kinematics}. A subset of these variable curvature shapes is shown in Fig. \ref{fig:planar_convergence_boxplot}. The shape of the rod was obtained as a solution to a boundary value problem using the shooting method for each applied wrench. The maximum applied force and moment were $\mb{f}_e = [\pm60,0,0] $ N and $\mb{m}_e = [0, \pm6, 0]\T$ Nm, with 10 wrenches selected between these maximum values, for a total of 100 applied wrenches and rod shapes that were solved for.
\par For each set of 100 shapes and number of strings considered, we determined the string routing radii and anchor points by solving the following constrained optimization problem:
\begin{equation} \label{eq:planar_optim_problem}
\max_{r_{x_i},s_{a_i}} \;  \aleph\left( \mb{J}_{\ell c}  \right) \quad \textrm{s.t.} \; \; 0 \leq s_{a_i} \leq L \quad i = 2,3,\dots, p
\end{equation}
where we have assumed that $r_{x_1} = 0.25$ and $s_{a_1} = L$ to represent an actuation tendon anchored at the end disk. We solved \eqref{eq:planar_optim_problem} using the interior point solver provided by MATLAB's \emph{fmincon()} with an initial guess of evenly spaced radii and anchor points. For each additional string, we used the Cosserat rod model to determine the string lengths, then used these simulated string lengths to predict the shape and tip pose $\mb{T}(L)$ by solving \eqref{eq:shape_sense_problem}.
\par We report the error between the tip pose predicted by the mechanics model and the solution given by solving \eqref{eq:shape_sense_problem} as a percent of the segment length:
\begin{equation}
e_p = \frac{\| \mb{p}_s - \mb{p}_p\|}{L} \times 100
\end{equation}
where $\mb{p}_s$ and $\mb{p}_p$ are the tip positions given by the mechanics model and predicted via \eqref{eq:shape_sense_problem}, respectively. The position errors across the 100 shapes and for different numbers of strings are shown in Fig. \ref{fig:planar_convergence_boxplot}. We do not report rotation errors because, as shown in \cite{simaan2004dexterous}, one string anchored to the end disk is sufficient to provide the tip angle of a planar segment and the rotation errors were therefore within numerical precision across all of the simulations.
\par Fig. \ref{fig:planar_convergence_boxplot} shows rapid convergence of the tip position error as the number of strings is increased. The average tip position errors across the 100 simulated shapes was 15\%, 0.47\%, 0.059\% and 0.0052\% of the segment length for one, two, three and four strings, respectively. We proceed with this case study by choosing $p = 3$.
\par Concatenating the string lengths for $i = [1,2,3]$ results in
\begin{equation} \label{eq:planar_string_lengths}
\bs{\ell} =  \begin{bmatrix}
s_{a_1} \\
s_{a_2} \\
s_{a_3} \\
\end{bmatrix}\underbrace{- \begin{bmatrix}
r_{x_1} & 0 & 0 \\
0 & r_{x_2} & 0 \\
0 & 0 & r_{x_3}
\end{bmatrix} \begin{bmatrix}
\int_0^{s_{a_1}}\bs{\phi}\T_y(s)\, ds \\[3pt]
\int_0^{s_{a_2}}\bs{\phi}\T_y(s)\, ds \\[3pt]
\int_0^{s_{a_3}}\bs{\phi}\T_y(s)\, ds \\[3pt]
\end{bmatrix}}_{\mb{J}_{\ell{}c}} \mb{c}_y
\end{equation}
where we have denoted the Jacobian $\mb{J}_{\ell{}c}$ with an underbrace. For the planar case, $\mb{J}_{\ell{}c}$ is independent of the configuration $\mb{c}$ (i.e. it is constant throughout the workspace).
\par Solving for the modal coefficients $\mb{c}_y$ requires inverting $\mb{J}_{\ell{}c}$. This Jacobian is the product of a diagonal matrix whose elements are the pitch radii of each string and a Vandermonde-like matrix of polynomial functions determined by the choice of modal basis. Here we seek to design the string routing paths to improve the numerical conditioning of $\mb{J}_{\ell{}c}$. For a given choice of the modal basis $\bs{\phi}_y$, the design parameters for each string are 1) the pitch radii $r_{x_i}$, and 2) the anchor points $s_{a_i}$.
\begin{figure*}[tbp]
  \centering
  \includegraphics[width=0.99\textwidth]{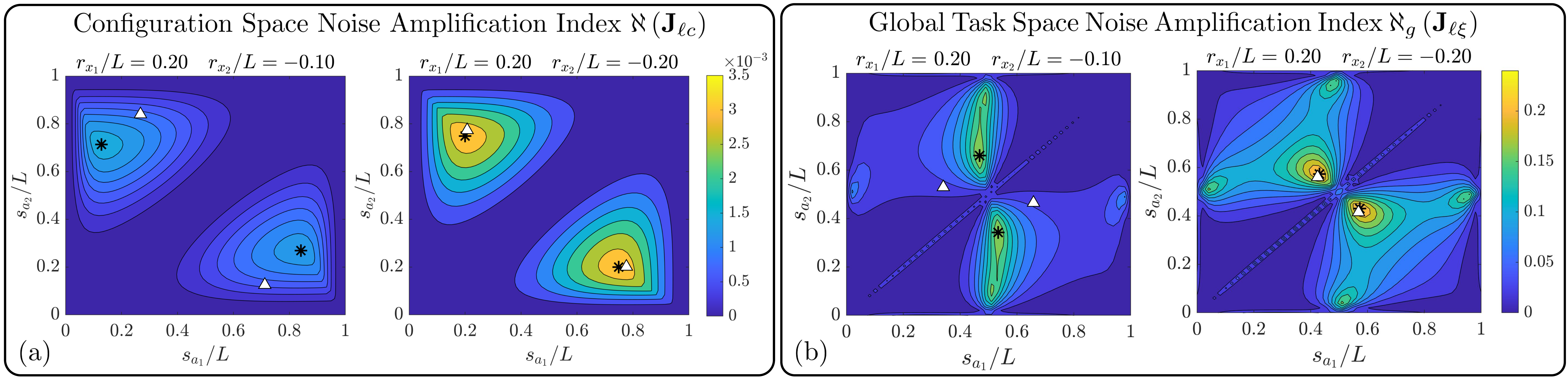}
  \caption{The values of a) $\aleph(\mb{J}_{\ell c})$ and b) $\aleph(\mb{J}_{\ell \xi})$ for different choices of string anchor points ($\frac{s_{a_1}}{L}$ and $\frac{s_{a_2}}{L}$) shown for two sample string radii ($\frac{r_{x_1}}{L}$ and $\frac{r_{x_2}}{L}$) in the planar robot shown in Fig.~\ref{fig:disk_collision}. The location of the two peaks on each plot are marked with an asterisk $(\ast)$, and a triangle $(\triangle)$ denotes the peak values for the other designs included in Table \ref{table:optimal_Jlc_planar} and Table \ref{table:optimal_Jellxi_planar}. The contours for the peaks denoted with triangles are not shown for clarity, but they follow a similar pattern to the ones shown.} \label{fig:planar_aleph}
\end{figure*}
\par There are several string routing design insights that can be observed directly from \eqref{eq:planar_string_lengths}. First, $r_{x_i}$ must not be zero to avoid rank deficiency. Second, increasing the pitch radius $r_{x_i}$ reduces the sensitivity of $\mb{c}_y$ to changes in $\bs{\ell}$, so increasing $r_{x_i}$ (while keeping the ratios of the pitch radii close to one) will increase $\aleph(\mb{J}_{\ell{}c})$. Third, if any two string anchor points $s_{a_i}$ are equal, $\mb{J}_{\ell{}c}$ will lose rank, so the strings should be anchored at unique $s_{a_i}$ along the segment. Noting that actuation tendons endowed with motor encoder sensing provide the same shape information as a passive string encoder, this means that although actuation tendons are commonly anchored at the end disk to expand the segment's workspace \cite{webster2010cc_review}, any additional actuation tendon anchored to the end disk provides no additional shape information (in the planar case). In Section \ref{sec:experiments_mfg}, we provide conditions under which strings or tendons anchored at the end disk do provide additional shape information for out-of-plane deflections.
\par We will now consider how to choose $s_{a_i}$ to maximize $\aleph(\mb{J}_{\ell c})$ (the noise amplification index of the configuration space Jacobian) for our planar example. To represent an actuation tendon that is anchored at the end disk, we choose $s_{a_3} = L$. We choose $r_{x_3} = 0.25L$ to correspond to the typical length-diameter ratio of most continuum robots. We now investigate how to optimally choose the radii and anchor points of the two remaining strings.
\par Figure \ref{fig:planar_aleph}a and Table \ref{table:optimal_Jlc_planar} show $\aleph(\mb{J}_{\ell c})$ for different choices of string radius and anchor point, from which we make several observations. First, we note that each plot in Fig. \ref{fig:planar_aleph}a contains two peaks, and that $\aleph(\mb{J}_{\ell c}) = 0$ when $s_{a_1} = s_{a_2}$ and when $s_{a_i} = 0$ or $s_{a_i}= L$. Second, we observe that increasing the pitch radius of both string increases $\aleph(\mb{J}_{\ell c})$ while not substantially changing the peak location. Third, we note that increasing a single string's pitch radius increases $\aleph(\mb{J}_{\ell c})$ while allowing the location of the peak to be adjusted.
\begin{table}[t] \centering
\caption{Planar Case: Local Maxima of $\aleph(\mb{J}_{\ell c})$ Compared to Equidistant Spacing of String Placement. $\beta$ Shows the Percent Improvement in $\aleph(\mb{J}_{\ell c})$ }\label{table:optimal_Jlc_planar}
\begin{tabular} {|c|c|c|c|c|c|} \hline
$r_{x_1}/L$ & $r_{x_2}/L$  & $s_{a_1}/L$ & $s_{a_2}/L$ & $\aleph(\mb{J}_{\ell c}) \; (\times 10^{-3})$ & $\beta$\\  \thickhline
\multirow{2}{*}{0.10} & \multirow{2}{*}{-0.10} & 0.204 & 0.772 & 1.03 & 64\%\\[-0.5pt]  \hhline{~~----}
&  & 0.772 & 0.204 & 1.03 & 64\%\\[-0.5pt]   \thickhline
\multirow{2}{*}{0.10} & \multirow{2}{*}{-0.20}  & 0.269 & 0.841 & 1.32 & 54\%\\[-0.5pt] \hhline{~~----}
&  & 0.714 &  0.128 & 1.50 & 76\%\\[-0.5pt]  \thickhline
\multirow{2}{*}{0.20} & \multirow{2}{*}{-0.10}  & 0.128 & 0.714 & 1.50 & 65\% \\[-0.5pt]  \hhline{~~----}
&  & 0.841 & 0.269 & 1.32 & 46\%\\[-0.5pt]   \thickhline
\multirow{2}{*}{0.20} & \multirow{2}{*}{-0.20}  & 0.200 & 0.749 & 3.29 & 53\%\\[-0.5pt]  \hhline{~~----}
&  & 0.749 & 0.200 & 3.29 & 53\%\\[-0.5pt]   \hline
\end{tabular}
\end{table}
\par Of note is the fact that the intuitive design choice of evenly spacing out the anchor points along $s = [0,L]$ does not result in the best kinematic conditioning for $\mb{J}_{\ell c}$. Table \ref{table:optimal_Jlc_planar} shows the values of the noise amplification index for each of the peaks in Fig. \ref{fig:planar_aleph} as well as for designs where the string anchor points are evenly spaced, i.e. $s_{a_1} = s_{a_2}$. Table \ref{table:optimal_Jlc_planar} also shows the improvement in the noise amplification index over the evenly spaced designs, calculated as:
\begin{equation}
\beta = \left[\left(\aleph\left(\mb{J}_{\ell c}\right) - \aleph\left(\widetilde{\mb{J}}_{\ell c}\right)\right)\middle/\aleph\left(\widetilde{\mb{J}}_{\ell c}\right)\right] \times 100
\end{equation}
where $\widetilde{\mb{J}}_{\ell c}$ denotes the Jacobians for designs having evenly spaced wire anchor points at $s_{a_1}=L/3, s_{a_2}=2L/3, s_{a_3}=L$. As shown in Table \ref{table:optimal_Jlc_planar}, placing the anchor points at one of the peaks instead of using evenly spaced anchor points resulted in 46\% or greater improvement in $\aleph\left(\mb{J}_{\ell c}\right)$ for all the cases considered.
\par We now consider improving the numerical conditioning of the full kinematic mapping (joint to task space) $\aleph_g\left( \mb{J}_{\ell \xi}(L) \right)$. We computed $\aleph_g\left( \mb{J}_{\ell \xi}(L) \right)$ for this robot using 67 configurations sampled in the admissible workspace, which we defined as any configuration that did not exceed 5\% strain for a 4 mm central backbone. Figure \ref{fig:planar_aleph}b shows the noise amplification for different string radii and anchor points. We observe that the design parameters that optimize $\aleph_g\left( \mb{J}_{\ell \xi}(L) \right)$ are not the same parameters that optimize $\aleph\left( \mb{J}_{\ell c} \right)$ in Fig. \ref{fig:planar_aleph}a, and in fact there is a conflict between these two design objectives, since the peaks in Fig. \ref{fig:planar_aleph}b correspond to valleys in Fig. \ref{fig:planar_aleph}a. Since the design objective is usually to minimize the pose error at a particular location, designing for maximizing  $\aleph_g\left( \mb{J}_{\ell \xi}(L) \right)$ is sufficient as long as the string routing solution avoids singularity of  $\mb{J}_{\ell c}$.
\begin{table}[t] \centering
\caption{Planar Case: Local Maxima of $\aleph_g(\mb{J}_{\ell \xi})$ Compared to Equidistant Spacing of String Placement. $\beta$ Shows the Percent Improvement in $\aleph(\mb{J}_{\ell \xi})$} \label{table:optimal_Jellxi_planar}
\begin{tabular} {|c|c|c|c|c|c|}\hline
$r_{x_1}/L$           & $r_{x_2}/L$            & $s_{a_1}/L$ & $s_{a_2}/L$ & $\aleph_g(\mb{J}_{\ell \xi}(L)) $ & $\beta$\\ \thickhline
\multirow{2}{*}{0.10} & \multirow{2}{*}{-0.10} & 0.573       & 0.428       & 0.159           & 146\% \\[-0.5pt]  \hhline{~~----} 
                      &                        & 0.428       & 0.573       & 0.159           & 146\% \\[-0.5pt]   \thickhline
\multirow{2}{*}{0.10} & \multirow{2}{*}{-0.20} & 0.343       & 0.534       & 0.181           & 204\% \\[-0.5pt]  \hhline{~~----} 
                      &                        & 0.660       & 0.468       & 0.181           & 204\% \\[-0.5pt]   \thickhline
\multirow{2}{*}{0.20} & \multirow{2}{*}{-0.10} & 0.468       & 0.660       & 0.181           & 213\% \\[-0.5pt]  \hhline{~~----} 
                      &                        & 0.534       & 0.343       & 0.181           & 213\% \\[-0.5pt]   \thickhline
\multirow{2}{*}{0.20} & \multirow{2}{*}{-0.20} & 0.573       & 0.431       & 0.228           & 76\% \\[-0.5pt]  \hhline{~~----} 
                      &                        & 0.431       & 0.573       & 0.228           & 76\% \\[-0.5pt]  \hline
\end{tabular}
\end{table}
\par We also observe again that the intuitive design of using evenly spaced anchor points does not result in an optimally conditioned kinematic mapping. Table \ref{table:optimal_Jellxi_planar} shows the values of $\aleph(\mb{J}_{\ell \xi}(L)) $ for the peaks shown in Fig. \ref{fig:planar_aleph}b as well as the percent improvement in $\aleph(\mb{J}_{\ell \xi}(L))$ as compared to a design using evenly spaced anchor points, i.e. $s_{a_2} = \frac{L}{3}$ and $s_{a_3} = \frac{2L}{3}$. The peak values of $\aleph_g(\mb{J}_{\ell \xi}(L)) $ were increased by 76\% or greater for all cases considered.
\par In this planar example, we have shown that increasing the number strings reduces the pose estimation error, provided string path design considerations, and showed that the intuitive choice of evenly spaced anchor points does not lead to optimal values for the noise amplification. The optimal placement of string anchor points depends to some degree on the family of deflected shapes that a given robot experiences under target design operating conditions (loading and reach). We have simulated the same case using a monomial basis as opposed to a Chebyshev polynomial basis and noted that the peaks in Fig.~\ref{fig:planar_aleph} experience negligible changes. Nevertheless, the designer should carry out a simulation as in Fig.~\ref{fig:planar_aleph} to achieve a qualitative understanding of the optimal location of wire anchor points and then determine these points while respecting practical design considerations. Building on these simulation results, we will now demonstrate our modeling and shape sensing approach on two other continuum robot embodiments that are subject to more realistic spatial deflections.
\section{Robots with High Torsional Stiffness and Constant Pitch String Paths} \label{sec:experiments_mfg}
\par We now consider robots that are subject to spatial deflections but have sufficiently high torsional stiffness that renders the torsional deflections negligible. We consider this category of robots because a number of continuum robots with high torsional stiffness have been presented in prior work \cite{dong2016twin_pivot,childs2020bellows,santoso2020origami,childs2021high_strength}, and we will present here a new modular collaborative continuum robot in this category and use it to experimentally validate the model we present. This model is also directly applicable to hyper-redundant robots with torsionally stiff universal joint backbones, e.g. \cite{anderson1970tensor_patent,hannan2003elephant}. This category of robots is also of interest because, as we will show, if the string paths are restricted to constant pitch radius paths, the configuration space Jacobian is constant and the modal coefficients are linear with respect to the string lengths, which simplifies the solution of the shape sensing problem.
\par First, we will present the kinematic formulation for this category of robots and provide considerations for designing the string paths. In particular, we show that
\emph{two strings anchored to the same disk add distinct information only if their polar coordinates are distinct by angle difference other than $0^\circ$ and $180^\circ$ (i.e. they are not collinear in the radial direction of the disk)}.
We then present the results of the string routing optimization for our collaborative continuum segment and experimentally validate the sensing approach, showing that the end disk position can be sensed with position errors below 5\% of arc length using information from four passive string encoders and two actuators.
\subsection{Kinematic model for torsionally stiff continuum robots} \label{subsec:experiments_mfg_kinmodel}
\par Under the assumption of negligible torsional deflections, i.e. $u_z(s) = 0$ the shape basis \eqref{eq:curvature_basis} takes the form
\begin{equation} \label{eq:curv_basis_zeros_torsion}
\mb{u}(s) =
\begin{bmatrix}
\bs{\phi}_x\T & 0  \\
0 & \bs{\phi}_y\T  \\
0 & 0
\end{bmatrix} \begin{bmatrix} \mb{c}_x \\ \mb{c}_y \end{bmatrix} = \bs{\Phi}(s)\mb{c}
\end{equation}
We restrict our consideration here to constant pitch-radius string routings give by:
\begin{equation} \label{eq:constant_pitch_path_mfg}
{}^t\mb{r}_i(s) = \begin{bmatrix} r_{x_i} & r_{y_i} & 0 \end{bmatrix}\T, \; r_{x_i} \in \realfield{}, \; r_{y_i} \in \realfield{}
\end{equation}
Noting that ${}^t\mb{r}'_i(s) = [0,0,0]\T$, the string path derivative \eqref{eq:string_path_deriv} simplifies to:
\begin{equation}
{}^t\mb{w}'_i(s) = \left[0, \,  0, \, \left(r_{y_i}\bs{\phi}_x\T\mb{c}_x - r_{x_i}\bs{\phi}_y\T\mb{c}_y + 1\right) \right]\T
\end{equation}
Applying the requirement from \eqref{eq:realizable_path_constraint} that $\left({}^t\mb{w}'_i\right)\T\mb{e}_3 > 0$ and using \eqref{eq:string_length} results in the string length as:
\begin{equation} \label{eq:zero_torsion_string_length}
\ell_i = s_{a_i} + \left(\int_{0}^{s_{a_i}} \begin{bmatrix} r_{y_i}\bs{\phi}_x\T, & -r_{x_i}\bs{\phi}_y\T \end{bmatrix} ds\right) \, \mb{c}
\end{equation}
where $\mb{c} = [\mb{c}_x\T, \mb{c}_y\T]\T$. Concatenating for each string $i \in [1,\dots,p]$ gives:
\begin{equation} \label{eq:zero_torsion_Jacobian}
\boldsymbol{\ell} = \begin{bmatrix}
s_{a_1} \\
\vdots \\
s_{a_p}
\end{bmatrix} +
\underbrace{\begin{bmatrix}
\int_{0}^{s_{a_1}} \begin{bmatrix} r_{y_1}\bs{\phi}_x\T, & -r_{x_1}\bs{\phi}_y\T \end{bmatrix} ds \\
\vdots \\
\int_{0}^{s_{a_p}} \begin{bmatrix} r_{y_p}\bs{\phi}_x\T, & -r_{x_p}\bs{\phi}_y\T \end{bmatrix} ds\\
\end{bmatrix}}_{\mb{J}_{\ell c}} \mb{c}
\end{equation}
As in the planar case, we observe that $\mb{J}_{\ell c}$ is independent of the configuration $\mb{c}$.
\par We now consider how to choose the anchor points and string paths to avoid singularities in $\mb{J}_{\ell c}$. Consider the scenario where two anchor points are equal, $s_{a_i} = s_{a_j}, i\neq j$. To prevent singularities, we must avoid a scenario where two rows of $\mb{J}_{\ell c}$ become dependent, e.g. one row is a scalar multiple of another. Considering a cross section of the segment at $s = s_{a_i} = s_{a_j}$, this will occur if the string radii $[r_{x_i}, r_{y_i}]$ and $[r_{x_j}, r_{y_j}]$ lie on a line passing through the origin of the body frame $\mb{T}(s_{a_i}) = \mb{T}(s_{a_j})$.
\par Furthermore, anchoring more than two strings will also cause $\mb{J}_{\ell c}$ to become rank-deficient. This is apparent since two radially non-collinear strings define the plane of the disk. Therefore, for an inextensible, torsionally stiff continuum segment, no more than two parallel-routed strings should be anchored to the same intermediate disk (to prevent rank-deficiency in $\mb{J}_{\ell c}$).
\subsection{Experimental validation on a collaborative continuum robot module}
\par
\begin{figure}[htbp]
  \centering
  \includegraphics[width=0.95\columnwidth]{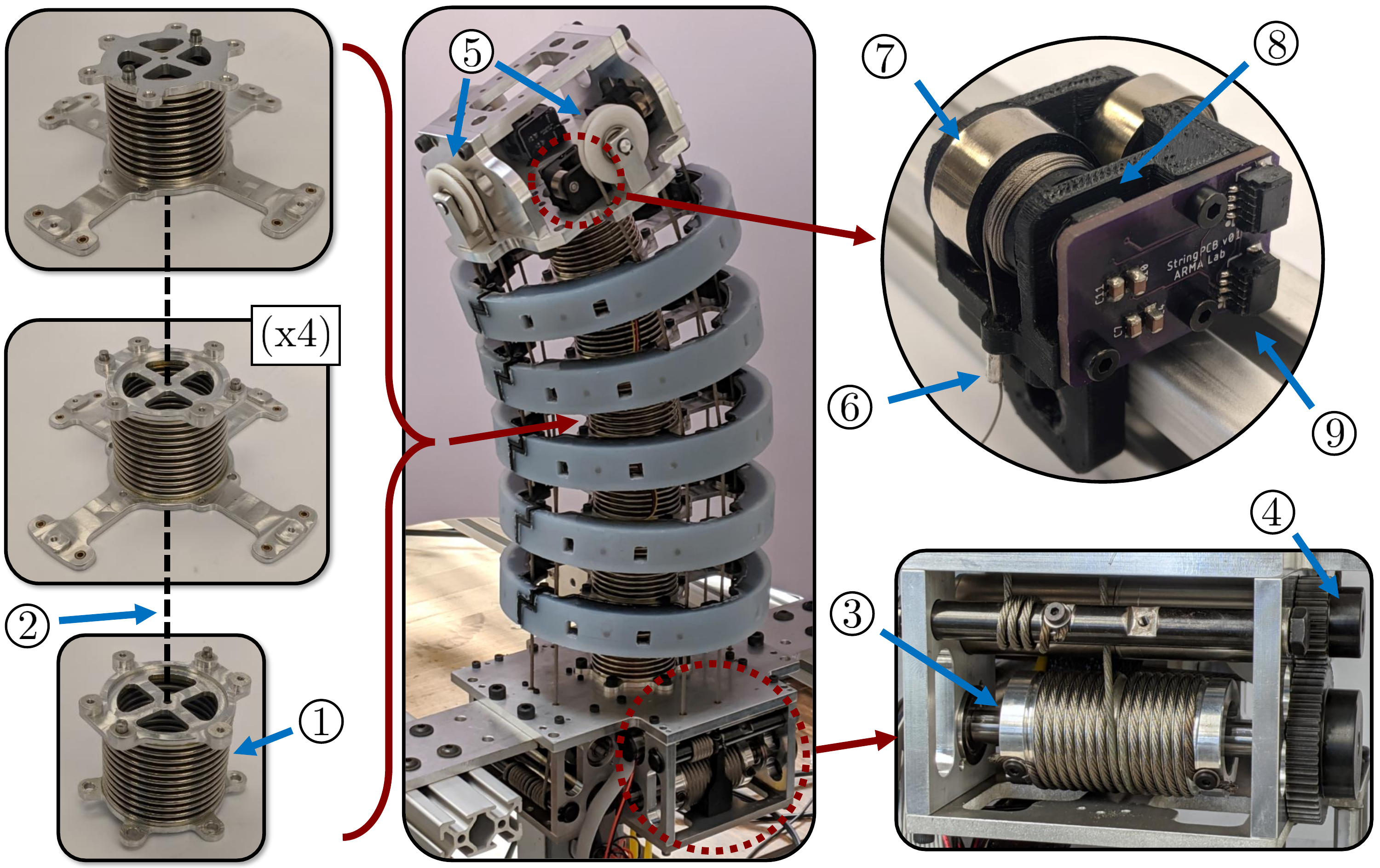}
  \caption{The modular continuum segment consists of torsionally stiff metal bellows  \protect\circled{1} with a \diameter 4 mm Nitinol rod \protect\circled{2} passing through their centers. The segment is actuated with a capstan mounted on a linear ball spline  \protect\circled{3} that is actuated by a brushed DC motor/geartrain \protect\circled{4}. Idler pulleys \protect\circled{5} provide a 2:1 reduction in the actuation tendon force. Each continuum segment contains four string encoders, which consist of a wire rope  \protect\circled{6}, a constant-torque return spring \protect\circled{7}, and a magnetic encoder \protect\circled{8} with an I$^2$C interface \protect\circled{9}. }
  \label{fig:mfg_segment}
\end{figure}

 We now validate our kinematic model on a continuum segment, shown in Fig. \ref{fig:mfg_segment} and in the multimedia extension, that is a portion of a robotic arm currently under development for collaborative manufacturing in confined spaces. Additional discussion on the motivations for development of this device can be found in \cite{abah2022sensors,johnston2020bracing}.
\par The flexible structure of the segment consists of aluminum intermediate disks with torsionally stiff metal bellows attached to the disks with an acrylic adhesive. Each bellow has a torsional stiffness of approximately 63 Nm/\textdegree, making the bellows approximately 1950 times stiffer in torsion than in bending. For the target application, the segments are expected to experience static torsional loads of up to 64 Nm, resulting in approximately  1\textdegree{} of torsional deflection per bellow. Since these maximum deflections are small relative to the bending deflections, we neglect the torsional deflections in our kinematic model below. These bellow subassemblies are bolted together, and a \diameter 4 mm solid Nitinol rod is passed through the center of the structure to prevent contraction of the segment. The length of the segment, measured from the top of the actuation unit to the bottom of the end disk, is 300.65 mm.
\par
\begin{figure}[htbp]
  \centering
  \includegraphics[width=0.99\columnwidth]{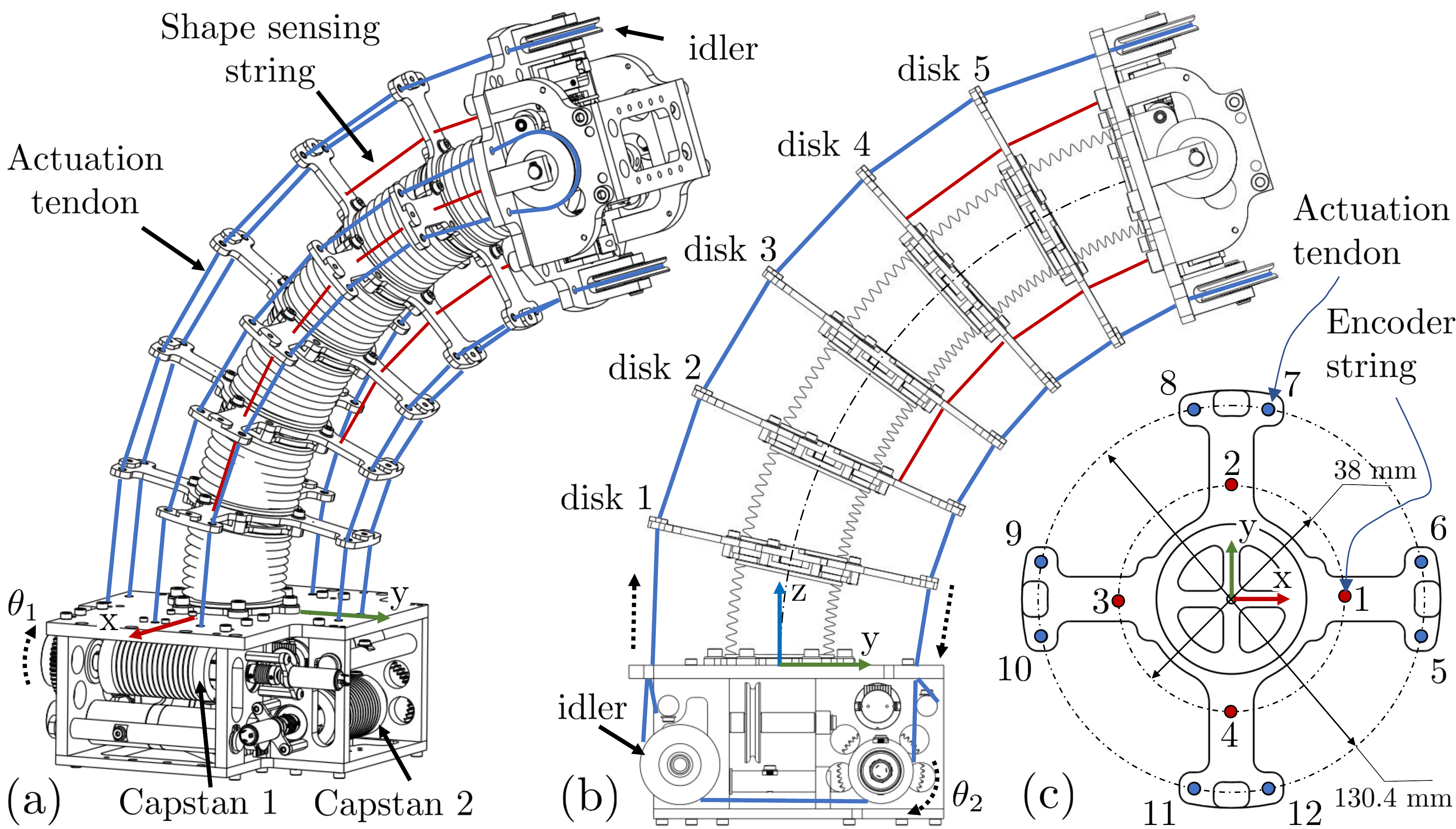}
  \caption{The path of the actuation tendons (numbered 5-12) and string encoders (numbered 1-4) are shown in (a) a 3D view, and (b) a simplified side view. Shown in (c) is top view of an intermediate disk with the locations where each string/tendon passes.}
  \label{fig:mfg_variables}
\end{figure}

 The segment is actuated by a pair of actuation tendons that pass through bronze bushings in the intermediate disks and are made from \diameter 2.38 mm steel wire rope. The actuation tendons are actuated in a differential manner (as shown in Fig. \ref{fig:mfg_variables}) by a capstan mounted on a linear ball spline, and this ball spline is actuated by a gearmotor with a 111:1 gear ratio (maxon DCX22L/GPX22HP) through a single spur gear stage with a gear ratio of 1.851. In the distal endplate assembly of the segment, idler pulleys route the tendons back towards the base of the segment, and the tendons are anchored to a manual pretensioning mechanism within the actuation unit. These idler pulleys provide a 2:1 reduction in the tendon force, which reduces the required sizing of the mechanical components in the actuation unit.
\par Each intermediate disk is also equipped with 8 time-of-flight proximity sensors and 8 contact sensors which enable mapping of the environment, human-robot physical interaction, and bracing against the environment along the entire body of the segment. The potential for whole-body contact with this segment is one motivation for the study of sensitivity analysis along different arc-lengths that we carry out below. Additional details on these sensing disks can be found in \cite{abah2019multimodal,abah2022sensors}.
\par Four custom-built string encoders are mounted in the distal portion of the segment, as shown in Fig. \ref{fig:mfg_segment}. The encoders have a \diameter{}0.33 mm diameter wire rope wrapped around a \diameter{}12 mm capstan and a constant-torque return spring (Vulcan Spring SV3D48) that results in a constant preload of 3.3 N on each string. This preload was sufficient to overcome friction and keep the string taught. Although the preload had a negligible effect on the robot shape, we note that our kinematic approach captures any deflections that occur due to the string preload. A printed circuit board (PCB) with a magnetic encoder (Renishaw AM4096 12 bit incremental) measures the angle of the capstan. The PCB also has connectors for the encoder's I$^2$C interface. All of these components are housed in a 3D-printed PLA housing. The magnetic encoder angle is read via I$^2$C by a Teensy 4.1 microcontroller at a rate of $\sim$150 Hz. The Teensy then provides the angle via UDP using a Wiznet WIZ850IO ethernet board to a Robot Operating System (ROS) node that publishes the string extensions on a ROS topic.
\par The segment motors are controlled with a control box containing six motor drivers (maxon ESCON 50/5), encoder reading cards (Sensoray 526) and a PC/104 CPU (Diamond Systems Aries) running Ubuntu/Linx with the PREEMPT-RT real-time patch. A higher-level ROS controller computes a fifth-order quintic polynomial trajectory and sends desired velocities to the motor control box via UDP. A real-time PI controller then commands desired motor currents based on the desired positon/velocity profile.
\par The radius of the capstan on each string encoder was calibrated by extending the string in 0.2 mm increments using a 3-DoF Cartesian robot. The linear stages of the Cartesian robot are Parker 404XR ballscrew linear stages actuated with brushed DC motors (Maxon RE35) and equipped with 1000 counts-per-turn encoders. The motion control accuracy of the linear stages was evaluated at $\pm 15$ $\mu$m. The Cartesian robot was used to extend/release the string over $n_c=1201$ positions given by  $\mb{d}_r=[0, 0.2, 0.4, \ldots,120, 119.8, 119.6, \ldots, 0]\T$ mm. Using the wrapping capstan model while neglecting the helix angle, we obtain the measurement model  $\mb{d}_r = \bs{\theta}_e r_e$, where $\bs{\theta}_e\in \realfield{n_c}$ is the vector of magnetic encoder angles and $r_e$ is radius of the encoder capstan. We then solve for the value of $r_e$ that minimizes the least-squares error between the predicted string extension based on string encoder angle and the distance traveled by the Cartesian robot by evaluating $r_e = \bs{\theta}_e^+ \mb{d}_r$. We calibrated all four of the string encoders and found that the average extension measurement error was below 0.1 mm (0.08\% of the total stroke) and the maximum extension measurement error across all four string encoders was 0.27 mm (0.23\% of total stroke).
\par Due to space considerations for integrating sensing electronics in the intermediate disks \cite{abah2022sensors}, the strings are restricted to be routed in constant pitch radius paths. The strings paths are therefore given by \eqref{eq:constant_pitch_path_mfg}, where the values for $r_{x_i}$ and $r_{y_i}$ are determined from the geometry in Fig. \ref{fig:mfg_variables}c. Since the string encoders are mounted in the distal endplate (instead of in the robot's base), the string lengths $\ell_i, \; i \in [1,2,3,4]$ are found via $\eqref{eq:string_length}$, except integration is performed from $s_{a_i}$ to $L$ rather than from $0$ to $s_{a_i}$, resulting in an expression similar to $\eqref{eq:zero_torsion_string_length}$.
\par We also use the actuation tendon lengths $\ell_i, \; i \in [5,\ldots, 12]$ as additional input to the problem of shape estimation. The routing path definitions for the actuation tendons require special consideration because of the idler pulleys in the end plate that reduce the tendon force by rerouting the tendons to the base of the robot. We separate the actuation tendon paths into eight different curves between the base plate and the end plate, as partially shown in Fig. \ref{fig:mfg_variables}a as blue curves. We then require the difference in the tendon path lengths on each side of the actuation capstan be equal to the change in length due to the actuation capstan rotation:
\begin{equation} \label{eq:mfg_tendon_lengths}
\begin{aligned}
\Delta\ell_{\theta_1}  &= \Delta\left(\ell_5 + \ell_6\right) =  -\Delta\left(\ell_9 + \ell_{10}\right) \\
\Delta\ell_{\theta_2} &= -\Delta\left(\ell_7 + \ell_8\right) = \Delta\left(\ell_{11} + \ell_{12}\right)  \\
\end{aligned}
\end{equation}
where $\ell_5 \ldots \ell_{12}$ refer to the actuated tendon lengths corresponding with the numbered bushings shown in Fig. \ref{fig:mfg_variables}c and $\ell_{\theta_1}$ and $\ell_{\theta_2}$ are the change in tendon length due to rotation of the capstan for joints 1 and 2, respectively. The values of $\ell_{\theta_1}$ and $\ell_{\theta_2}$ are determined from the motor angles while accounting for the helical wrapping pattern on the capstan:
\begin{equation}
\ell_{\theta_i} = \frac{\theta_i}{2\pi}\sqrt{\left( 2 \pi r_c \right)^2 + \gamma^2}, \quad i \in [1,2]
\end{equation}
where $\theta_1$ and $\theta_2$ are the angles of the first and second actuation capstans, respectively, $r_c$ is the radius of the capstan, and $\gamma$ is the lead of the helical groove on the capstan. The actuation tendon lengths $\ell_i, \; i \in [5 \dots 12]$ are found via \eqref{eq:zero_torsion_string_length}.
\par Information from the two motor encoders and the four string encoders allows for a modal basis with six columns ($p = 6$). Neglecting torsional deflections, the curvature distribution is given by \eqref{eq:curv_basis_zeros_torsion} with the following shape functions:
\begin{equation}
\bs{\phi}_x(s) = \bs{\phi}_y(s) = \begin{bmatrix} T_0, T_1(s), T_2(s) \end{bmatrix}\T\\
\end{equation}
The string encoder length equations for $i \in [1,2,3,4]$ are concatenated together with \eqref{eq:mfg_tendon_lengths} to give the string lengths in a similar form as in \eqref{eq:zero_torsion_Jacobian}, but accounting for the string encoder routing as described above while adding the tendon lengths:
\begin{equation} \label{eq:string_lengths_mfg}
\begin{gathered}
\begin{bmatrix}
\ell_1 \\ \ell_2 \\ \ell_3 \\ \ell_4 \\ \ell_{\theta_1} \\ \ell_{\theta_2}
\end{bmatrix} = \begin{bmatrix}
L - s_{a_1} \\ L - s_{a_2} \\ L - s_{a_3} \\ L -  s_{a_4} \\ 0 \\ 0
\end{bmatrix} + \mb{J}_{\ell c}\mb{c}, \quad  \mb{J}_{\ell c} \in \realfield{6 \times 6 }, \quad \mb{c} \in \realfield{6}
\end{gathered}
\end{equation}
\par Given a set of string measurements, we then solve \eqref{eq:string_lengths_mfg} for the modal coefficients $\mb{c}$ with a single matrix inversion. Our MATLAB 2019b implementation computes $\mb{c}$ at a rate of \mbox{$\sim 125$} kHz. As with the planar example above, the configuration space Jacobian $\mb{J}_{\ell c}$ is constant when assuming zero torsional deflections, so $\mb{J}_{\ell c}$ and $\mb{J}_{\ell c}^{-1}$ can be computed once and stored.
\par Based on the analysis in Section \ref{subsec:experiments_mfg_kinmodel} of singularities in $\mb{J}_{\ell c}$, the two actuation tendon equations in \eqref{eq:mfg_tendon_lengths} do not introduce singularities into $\mb{J}_{\ell c}$, because the tendon pairs that are collinear with the central backbone point are in the same equations in $\eqref{eq:mfg_tendon_lengths}$. However, any additional string anchored at the end disk (or equivalently at the base disk for the strings mounted in the endplate) would not provide additional shape information. The radius of the passive sensing strings is fixed, so in designing the string paths for this segment, we can only change the string anchor points. The anchor points are also restricted to the discrete points along the central backbone where the five intermediate disks lie. We therefore have five possible choices for $s_{a_i}$ for each of the four strings.
\par To design the string routings, we run a brute-force search across all possible combinations of $s_{a_i}$ (625 possible designs) to find the combination with the largest $\aleph_g \left( \mb{J}_{\ell \xi}(s) \right)$. We chose the characteristic length to be 0.0652, the kinematic radius of the actuation tendons. The global noise amplification index $\aleph_g$ was computed for a set of 320 configurations in the segment's admissible workspace, and the noise amplification index for the end disk, denoted as $\aleph_g \left( \mb{J}_{\ell \xi}(L) \right)$, and the noise amplification index for the third disk, denoted as $\aleph_g \left( \mb{J}_{\ell \xi}(s_3) \right)$, where $s_3$ is the arc length at which the third disk is located, were computed at each configuration.
\begin{figure}[htbp]
  \centering
  \includegraphics[width=0.95\columnwidth]{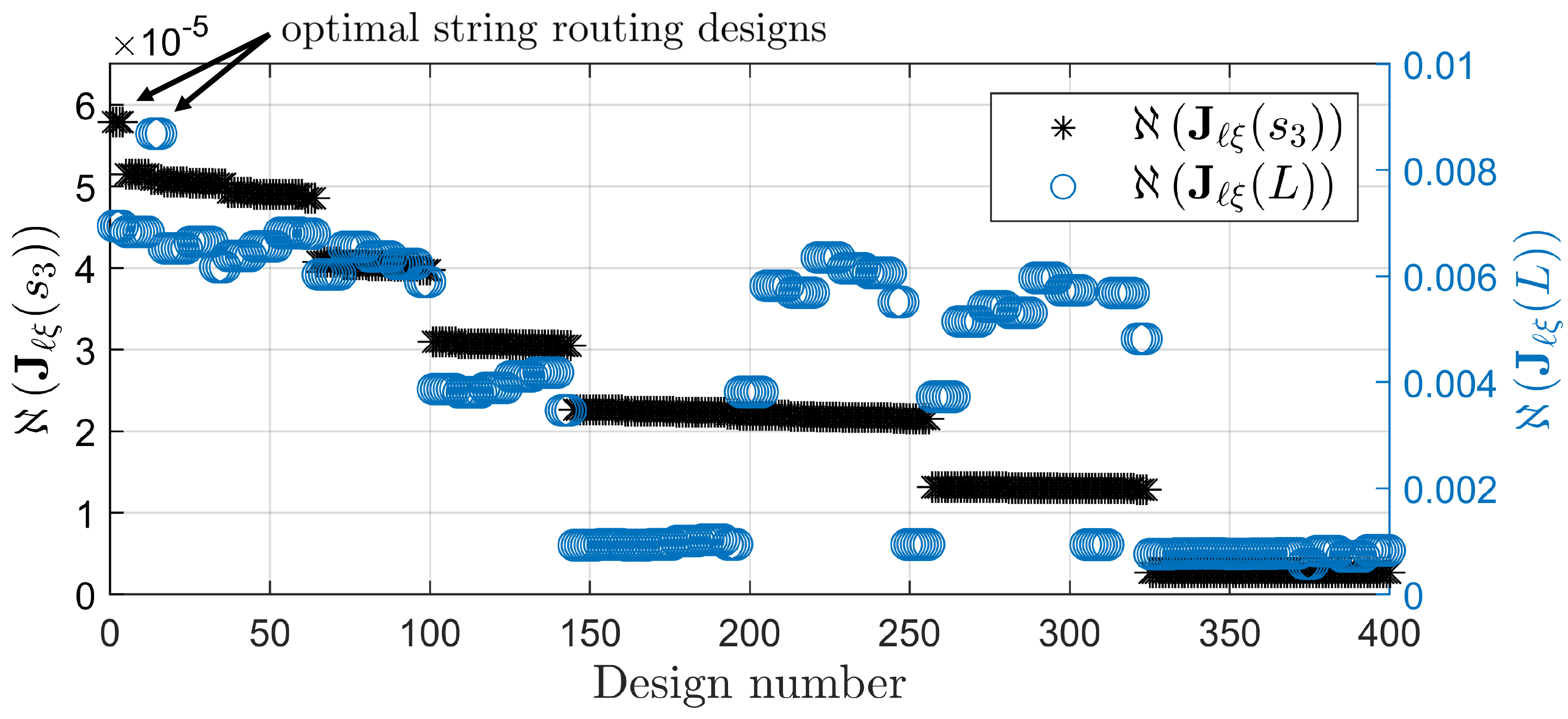}
  \caption{Noise amplification indices at the end disk and the third disk for all physically realizable string routing designs of the collaborative continuum robot. A large range of noise amplification indices are possible, but the noise amplification indices for the design optimal for end disk pose estimation is not significantly different than for the design optimal for Disk 3 pose estimation.}
  \label{fig:mfg_aleph_designs}
\end{figure}
\par Out of the 625 string routing designs considered, 225 resulted in a singular $\mb{J}_{\ell \xi}(L)$ and $\left( \mb{J}_{\ell \xi}(s_3)\right)$. Figure \ref{fig:mfg_aleph_designs} shows the noise amplification indices for 400 of the string routing designs in our brute-force search that did not result in singularities. Some designs resulted in the noise amplification being particularly close to zero. For example, one poorly conditioned routing design was a design with anchor points at disk 2, disk 4, disk 3, and disk 5, for strings 1-4, respectively, with the disk numbers given in Fig. \ref{fig:mfg_variables}. This design places one string on each disk except disk 1. Although this choice might seem reasonable to a designer at first glance, its noise amplification index is $\aleph_g ( \mb{J}_{\ell \xi}(L)) = 8.3\text{e-3}$, which is two orders of magnitude smaller than for the optimal design. This highlights the importance of carrying out the sensitivity analysis we present herein when choosing a string routing design to avoid these ill-conditioned string routings.
\par From the designs in Fig. \ref{fig:mfg_aleph_designs}, we found that the anchor points that maximize $\aleph_g \left( \mb{J}_{\ell \xi}(L) \right)$ were disk 4, disk 4, disk 2, and disk 2, for strings 1-4 respectively. Below, we will refer to this string routing design as the \emph{end disk routing}. For maximizing $\aleph_g \left( \mb{J}_{\ell \xi}(s_3) \right)$, the optimal anchor points were disk 1, disk 1, disk 3, and disk 3, for strings 1-4, respectively. Below, we will refer to this string routing design as the \emph{third disk routing}.
\begin{table}[ht] \centering
\caption{Noise Amplification Indices for Optimized String Routing Designs on Collaborative Continuum Module}{} \label{table:mfg_routing_aleph}
\fontsize{8}{10}\selectfont \renewcommand {\arraystretch} {1.15}
\begin{tabular} {|c|>{\centering\arraybackslash}p{18mm}|c|} \hline
& \multicolumn{2}{c|}{Disk Used For Routing Optimization}\\ \cline{2-3}
& End & 3rd \\ \hline
$\aleph_g \left( \mb{J}_{\ell \xi}(L) \right)$  & 8.69\text{e-3}  & 6.94\text{e-3}   \\ \hline
$\aleph_g \left( \mb{J}_{\ell \xi}(s_3) \right)$  &   5.08\text{e-5}  & 5.79\text{e-5} \\ \hline
\end{tabular}
\end{table}
\par The noise amplification indices for these two optimized designs are given in Table \ref{table:mfg_routing_aleph}. Although the brute-force search optimization increased $\aleph_g \left( \mb{J}_{\ell \xi}(L) \right)$ for the end disk routing by 25\% compared to the middle disk routing, and the value of $\aleph_g \left( \mb{J}_{\ell \xi}(L) \right)$ for the middle disk routing was increased by 14\% compared to the end disk routing, we will show in our experimental results below that these changes in $\aleph_g \left( \mb{J}_{\ell \xi}(L) \right)$ are not large enough to have a significant effect on the pose error either at the end disk or at the third disk. Either of these string routing designs could be chosen for this robot without having a significant effect on the pose error.
\begin{figure}[tbp]
  \centering
  \includegraphics[width=0.90\columnwidth]{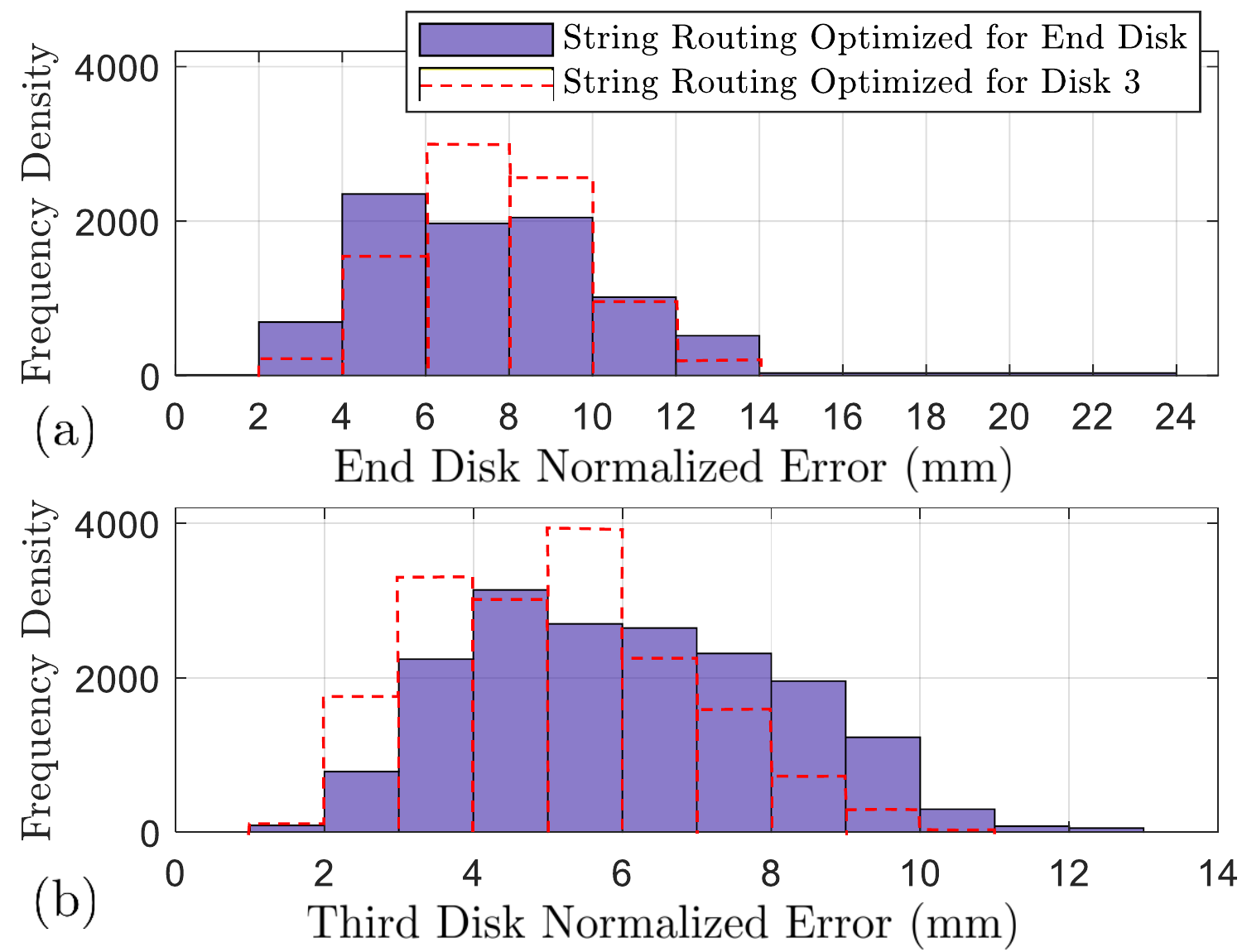}
  \caption{The frequency histogram of pose error when the string routing is optimized to minimize the effect of measurement noise for either the end disk pose (end disk routing) or the pose of the third disk (third disk routing). (a) The end-disk position error histograms for both routings showing that the error distribution for $\mb{T}(L)$ is shifted leftward compared to the third disk routing (b) The third-disk position error histograms for both routings showing that the error distribution for $\mb{T}(s_{3})$  is shifted leftward compared to the end disk routing.}
  \label{fig:mfg_histo}
\end{figure}
\par
\begin{figure}[htbp]
  \centering
  \includegraphics[width=0.75\columnwidth]{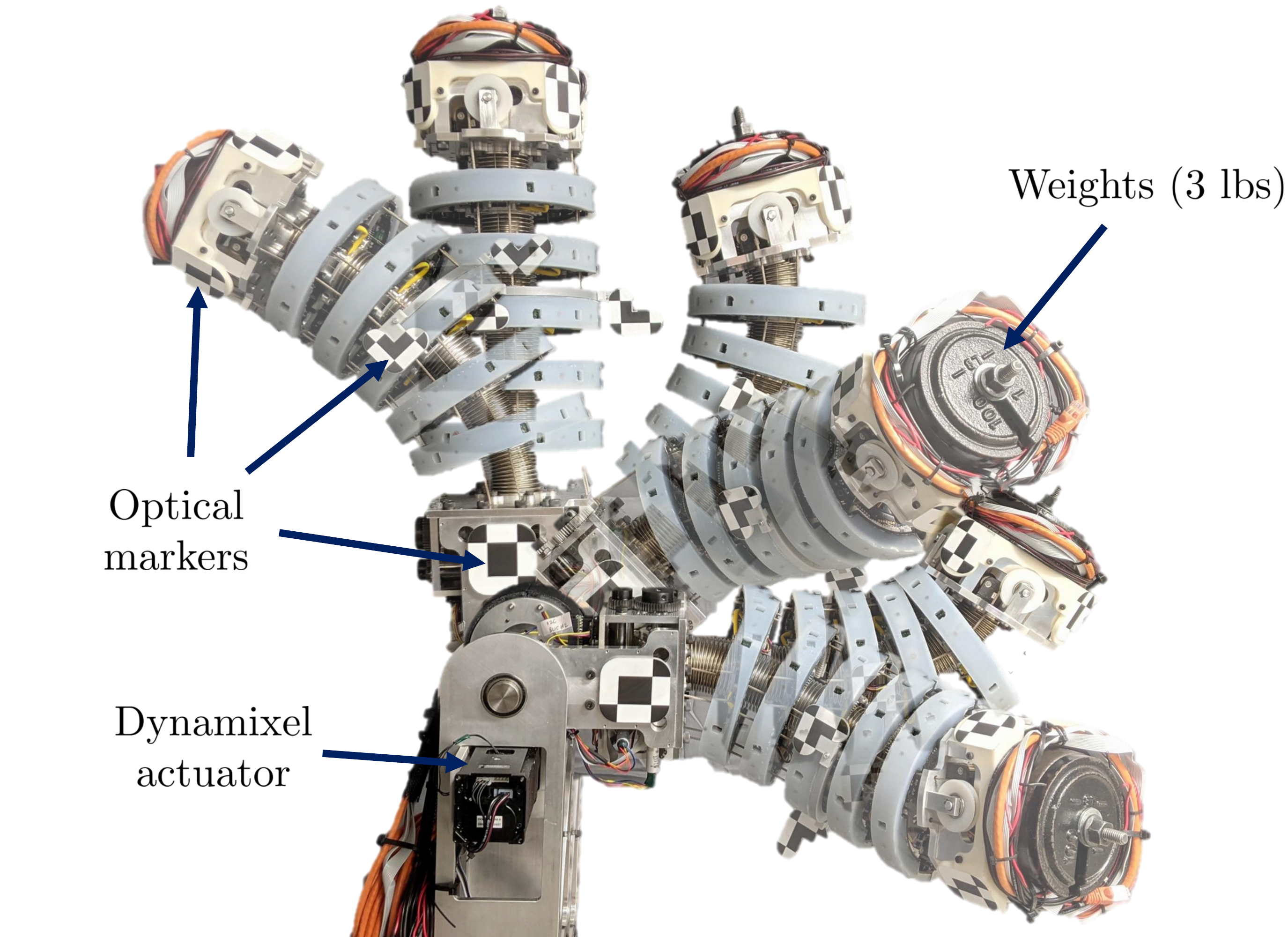}
  \caption{A subset of the variable curvature spatial configurations used to validate our shape sensing approach for a torsionally stiff continuum segment with straight string routing. The third and end disk poses were captured using optical trackers with and without weights attached to the end disk.}
  \label{fig:mfg_configs}
\end{figure}

 We experimentally validated the shape sensing approach on the physical continuum robot. We routed the string encoders according to the two routing designs given by our optimization procedure, and for each string routing design, we measured the shape of the segment across a large variety of variable curvature shapes, a subset of which are shown in Fig. \ref{fig:mfg_configs}. The segment was mounted on a revolute joint driven by an off-the-shelf actuator (Dynamixel PH54-200-S500-R), and the angle of the revolute joint was commanded to 0, 45\textdegree, and 90\textdegree. For each of these three angles, we commanded the segment to move from the initial home configuration to four different configurations: $\bs{\theta} = [0\degree, 500\degree]$, $\bs{\theta} = [0\degree, -500\degree]$, $\bs{\theta} = [500\degree, 0\degree]$, and $\bs{\theta} = [-500\degree, 0\degree]$, where $\bs{\theta} = [\theta_1, \theta_2]\T$ contains the angles of the actuation capstans. The motion profile to reach these four poses was generated using a fifth-order polynomial trajectory planner, with a time of 45 seconds to move to the desired configuration from the initial home configuration. During these motions, we continuously captured the string lengths using the \emph{rosbag} ROS package. We captured ground-truth pose of the third intermediate disk and the end disk using a stereo vision optical tracker (ClaroNav H3-60) and optical markers mounted on the robot. The data collection rate is limited by the optical tracker's $\sim{}8.5$ Hz sample rate. This experiment was carried out with the end disk routing, and then repeated for the third disk routing.
\par Using the string length measurements acquired with the string encoders, we solved for the modal coefficients using the full shape sensing model given by \eqref{eq:string_lengths_mfg}. To compare against a scenario where the robot did not have shape sensing encoders, we also reconstructed the shape using only the actuation variables $\bs{\theta}$. For this case, we used a modal basis with two columns where $\phi_x = \phi_y = 1$. This is identical to the commonly used constant-curvature model \cite{webster2010cc_review}. We report the constant-curvature model results using the data set collected with the third disk routing, but note that the errors were similar for this constant-curvature model using the end disk routing.
\begin{table}[ht] \centering
\caption{Average (Maximal) Absolute Position and Orientation Errors in mm and $^\circ$ for the Segment in Fig.~\ref{fig:mfg_segment}. The errors are Specified for the End Disk $(s=L)$ and the $3^{rd}$ Disk $(s=s_3)$}{} \label{table:mfg_errors}
\fontsize{8}{10}\selectfont \renewcommand {\arraystretch} {1.15}
\begin{tabular} {|>{\centering\arraybackslash}p{26mm}|>{\centering\arraybackslash}p{13mm}|>{\centering\arraybackslash}p{14mm}|>{\centering\arraybackslash}p{14mm}|} \hline
                                                           & End Disk Routing & Third Disk Routing & w/o Passive Strings\\ \thickhline
$\bar{p}_{e}(L)$, $\text{max}\left(p_{e}(L)\right)$               & 5.9 (14.4)   &  6.0 (13.8)       & 56.2 (104.9)    \\ \hline
$\bar{p}_{e}(s_3)$, $\text{max}\left(p_{e}(s_3)\right)$           & 3.8 (10.2)   &  3.3 (9.2)        & 31.8 (58.6)      \\ \hline
$\bar{\theta}_{e}(L)$, $\text{max}\left(\theta_{e}(L)\right)$     & 1.5 (8.6)    &  1.5 (3.9)        & 3.6  (14.4)       \\ \hline
$\bar{\theta}_{e}(s_3)$, $\text{max}\left(\theta_{e}(s_3)\right)$ & 2.0 (6.0)    &  1.6 (4.2)        & 15.4 (28.1)     \\ \hline
\end{tabular}
\end{table}
\par The mean and maximum errors of our shape sensing model (with the two different routing designs) as well as the constant curvature model are given in Table \ref{table:mfg_errors}. The position error is given by $p_e(s) = \| \mb{p}_{model}(s) - \mb{p}_{meas}(s) \|$, where $\mb{p}_{model}(s)\in \realfield{3}$ is the model-predicted position, and $\mb{p}_{meas}(s)\in \realfield{3}$ is the measured position, and we denote the average of $p_e(s)$ across all configurations as $\bar{p}_{e}(s)$. The angular error is given by:
\begin{equation} \label{eq:angular_error}
\theta_{e}(s) = \text{cos}^{-1}\left( \frac{\text{trace}(\mb{R}_{meas}(s)\mb{R}_{model}(s)\T - 1}{2} \right)
\end{equation}
where $\mb{R}_{meas}(s) \in SO(3)$ is the measured rotation matrix and $\mb{R}_{model}(s) \in SO(3)$ is the model-predicted rotation matrix. Both the end disk routing and the middle disk routing reduced the mean end disk errors compared to the constant curvature model by more than 89\% in position and 58\% in angle. The maximum end disk error was reduced compared to the constant curvature model by more than 85\% in position and 40\% in angle for both string routing designs. Both routing designs have average tip position errors below 2.0\% of the total arc length, a significant improvement compared to the constant curvature model error of 18.7\% of total arc length. The maximum tip position error of our approach was 4.8\% of the total arc length, compared to 34.9\% for the constant-curvature model.
\par The third disk routing reduced the mean position error at $s_3$ by 0.5 mm and the maximum by 1 mm. The third disk routing design also reduced the mean angular error at $s_3$ by 0.4\textdegree and the maximum by 1.8\textdegree. Figure \ref{fig:mfg_histo} shows the histograms of the normalized error $e_n \in \realfield{}$ at the end disk and the third disk:
\begin{equation} \label{eq:normalized_error}
e_n = \sqrt{\| \mb{p}_{model} - \mb{p}_{meas} \| + c_\ell\theta_{e}}
\end{equation}
where $c_\ell$ is the characteristic length used in the string routing design optimization procedure (in this case, we used the kinematic radius of the disk 0.1304/2 m as shown in Fig.~\ref{fig:mfg_variables}). We observe that, as expected, the error distribution of $\mb{T}(s_3)$ is shifted leftward towards reduced error due to the increase in $\aleph_g \left( \mb{J}_{\ell \xi}(s_3) \right)$. However, we also observe that the maximum errors for $\mb{T}(L)$ were higher using the routing optimized for the end disk. This indicates that the change in the noise amplification index at $\mb{T}(L)$ when using the routing optimized for the third disk was not significant enough to affect the tip pose error in a way that would overcome other sources of error, i.e. friction, mechanical clearances, and the continuously parallel routing assumption. Overall, while the general trends in the errors match the expected behavior due to the noise amplification index, we observe that the pose error of the middle and end disk was not substantially affected, meaning that either of the two optimized string routing designs could be used. This provides flexibility in the string routing design.
\par In this section, we have presented the kinematic formulation for sensing deflections of robots with negligible torsional stiffness. The formulation results in a constant configuration space Jacobian and linear shape sensing equations. We validated the model and approach for a collaborative continuum robot with high torsional stiffness and constant pitch radius string paths, showing that our approach sensed the tip position with errors below 4.8\% of total arc length. We now demonstrate the approach for the more general case of robots with non-negligible torsional deflections and helical string routing.
\section{Sensing Torsional Deflections with \\ Helical String Paths} \label{sec:experiments_soft}
\par In the section above, we validated our shape sensing approach on a segment with high torsional stiffness. Neglecting torsional stiffness significantly simplifies the model equations and reduces the computation cost of solving for the shape. However, many continuum robots have relatively low torsional stiffness, so torsional deflections cannot always be neglected. In this section, we consider helical string routing as a way to sense torsional deflections. We then validate the approach in a simulation study using a Cosserat rod mechanics model.
\begin{figure}[htbp]
  \centering
  \includegraphics[width=0.95\columnwidth]{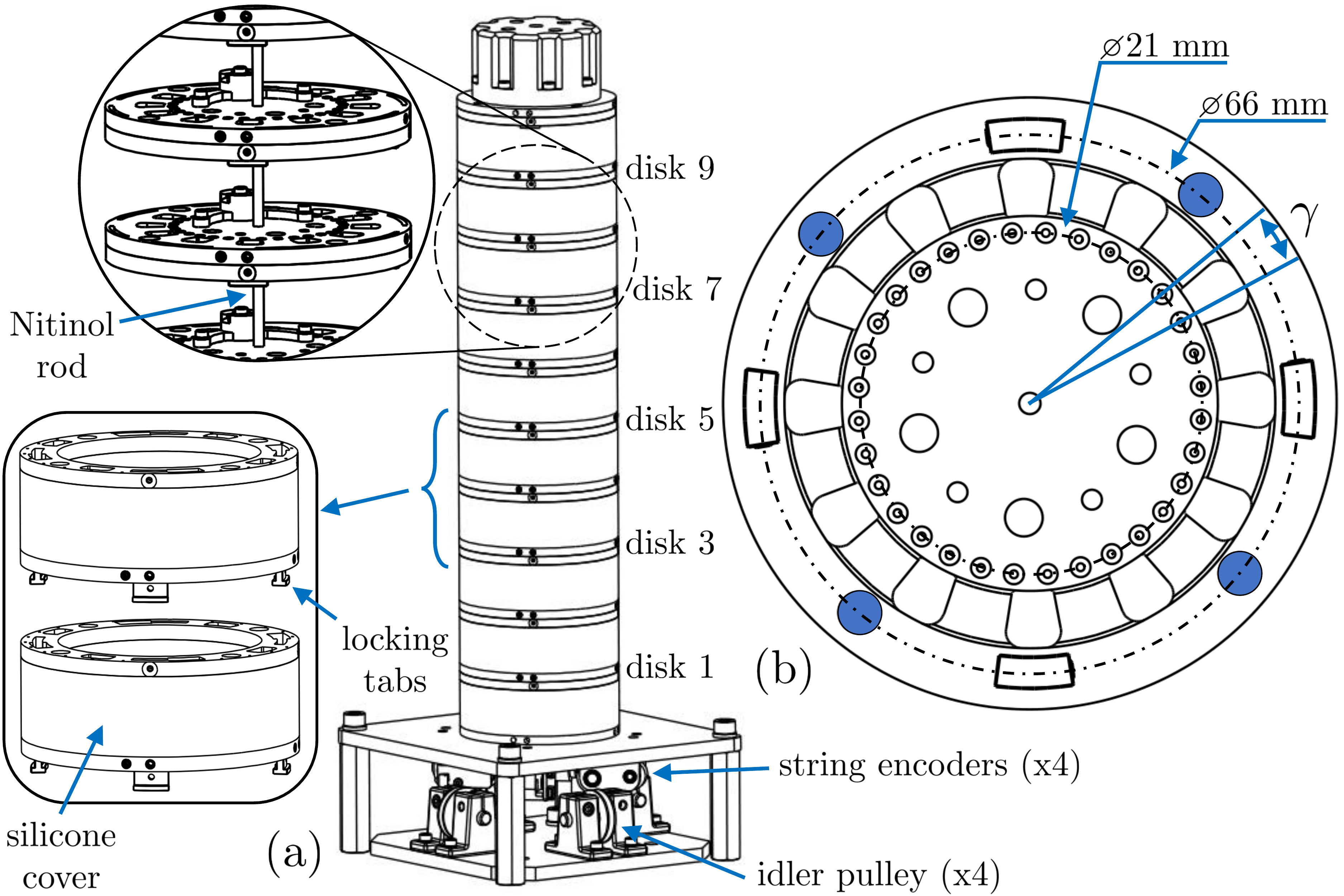}
  \caption{To validate our approach for torsional deflections and helical string routing, we simulated a modular soft continuum segment. The segment's interlocking subsegments are over-molded with silicone, and a Nitinol rod passes through their centers. Four string encoders are mounted at the segment's base, and each intermediate disk (b) has 32 holes to enable helical string paths.}
  \label{fig:bandybot}
\end{figure}
\par We chose the geometry of the simulated continuum robot based on a concept for a modular soft continuum robot, shown in Fig. \ref{fig:bandybot}. The subsegments of the robot are built with cylindrical silicone over-molded on the outer circumference of the intermediate disks to act as a soft outer cover, and the bottom plate of each subsegment has locking tabs that mate with a slot in the top plate of the previous subsegment. A 293 mm long solid Nitinol rod passes through the center of the continuum robot to prevent compression of the structure.
\par Four string encoders (as described in Section \ref{sec:experiments_mfg}) are mounted below the segment, and four tendons are routed to actuators via idler pulleys below the segment's base. In this embodiment, the actuation tendons are routed in straight paths with the tendon path given by \eqref{eq:constant_pitch_path_mfg}. We assume that two of the tendons are anchored at the end disk, and that two of the tendons are anchored at disk 7. This choice in anchor points for the actuation tendons allows all four actuation tendons to provide shape information (since more than two strings anchored to a disk will result in a singular $\mathbf{J}_{\ell c}$ when the robot is straight) while still allowing the segment to bend in all four directions and have a large reachable workspace. The string encoders are routed in helical paths given by \eqref{eq:helical_path}. As shown in Fig. \ref{fig:bandybot}c, the bottom plate of each subsegment has 32 holes through which the strings pass, allowing the strings to be routed in the desired helical shape. We now optimize the anchor points of the four string encoders and the twist rate of their helical paths.
\par The helical routing string path function is given by \cite{rucker2011statics}:
\begin{equation} \label{eq:helical_path}
{}^t\mb{r}_i(s) = r_s[\text{cos}(\omega{}s + \alpha_i),\text{sin}(\omega{}s + \alpha_i),0]\T
\end{equation}
where $r_s$ is the radius of the helical path, $\omega$ is the twist rate of the helical path (which we assumed was constant and equal for all four strings to prevent the string paths from intersecting), and $\alpha_i$ is an angular offset for each string. Since we have eight inputs to our shape sensing model (four passive string encoders and four active actuation tendons), we choose a modal basis with eight columns. The modal shape basis is given by \eqref{eq:curvature_basis} with the following shape functions:
\begin{equation} \label{eq:bb_modal_funcs}
\begin{gathered}
\bs{\phi}_x(s) = \bs{\phi}_y(s) = \begin{bmatrix} T_0, T_1(s), T_2(s) \end{bmatrix}\T\\
\bs{\phi}_z(s) = \begin{bmatrix} T_0, T_1(s) \end{bmatrix}\T
\end{gathered}
\end{equation}
where the Chebyshev functions $T_i(s)$ are given by \eqref{eq:cheby_first_three}. We compute the string lengths and the configuration space Jacobian $\mb{J}_{\ell c}$ by numerically integrating \eqref{eq:string_length} and \eqref{eq:config_jacobian_row}, respectively.
\par We solved the optimization problem \eqref{eq:design_problem} with $\epsilon = 1\text{e-7}$ through a brute-force search of all possible string anchor points (disks 1-10) and helical path design parameters. Given the 32 holes on the intermediate disks, as shown in Fig. \ref{fig:bandybot}, the helical path twist rate can be approximated as $\omega = n_{\omega}\gamma$, where $n_{\omega} \in \realfield{}$ determines the number of routing holes to skip in between the intermediate disks when routing the string, and $\gamma = \frac{2\pi}{32}$ is the angle between the routing holes, as shown in Fig. \ref{fig:bandybot}. We constrained the twist rate to $n_{\omega}<2$ to prevent large twist rates, since excessive twist rates increase the possibility of binding between the string and the routing holes. With the ten possible string anchor points and two possible twist rates for four strings, there were 20,000 possible string routing designs that we evaluated. Using a prototype subsegment, we experimentally determined that each subsegment can bend up to 10\textdegree{} and twist up to 7.5\textdegree{} without mechanical failure, so the admissible workspace was defined as a set of 32 sampled configurations that did not exceed 10\textdegree{} of bending and 7.5\textdegree{} of twist in each subsegment. We chose the characteristic length to be radius of the segment, 37.5 mm. We discarded all designs that violated $\aleph_g(\mb{J}_{\ell c}) \geq 1\text{e-7}$. We will consider below the fourth, sixth, and end disk as representative examples to optimize $\mb{J}_{\ell \xi}$ for, so we stored the noise amplification indices for $s=L$, $s=s_4$ (the location of the fourth disk), and $s = s_6$ (the location of the sixth disk).
\begin{figure}[htbp]
  \centering
  \includegraphics[width=0.95\columnwidth]{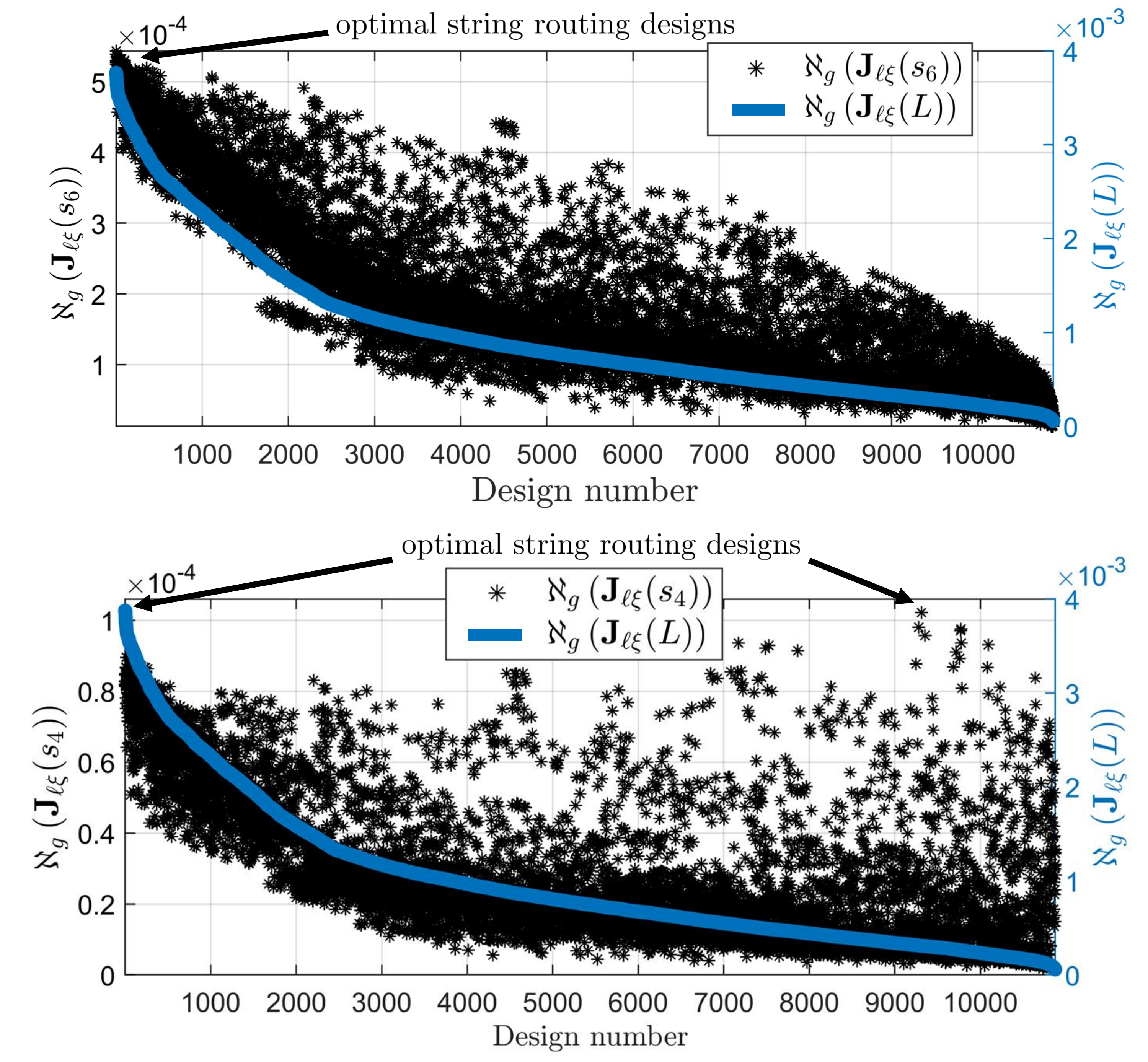}
  \caption{The noise amplification indices across all physically realizable string routing designs for the simulated soft robot with torsional deflections. The routing that maximizes $\aleph_g(\mb{J}_{\ell \xi}(s_4))$ results in a significantly reduced $\aleph_g(\mb{J}_{\ell \xi}(L))$, but for disk 6, $\aleph_g(\mb{J}_{\ell \xi}(L))$ does not significantly change.}
    \label{fig:bb_aleph_designs}
\end{figure}
\begin{table}[ht] \centering
\caption{Noise Amplification Indices for Optimized String Routing Designs on Simulated Segment with Helical Routing} \label{table:bb_routing_aleph} \renewcommand {\arraystretch} {1.2}
\begin{tabular} {|c|c|c|c|} \hline
& \multicolumn{3}{c|}{Disk Used For Routing Optimization} \\ \cline{2-4}
                                                 & End            & 6th            & 4th             \\ \hline
$\aleph_g \left( \mb{J}_{\ell \xi}(L) \right)$   & 3.47\text{e-3} & 3.50\text{e-3} & 3.35\text{e-4}  \\ \hline
$\aleph_g \left( \mb{J}_{\ell \xi}(s_6) \right)$ & 5.24\text{e-4} & 5.40\text{e-4} & 2.27\text{e-4}  \\ \hline
$\aleph_g \left( \mb{J}_{\ell \xi}(s_4) \right)$ & 7.91\text{e-5} & 8.27\text{e-5} & 8.85\text{e-5}  \\ \hline
\end{tabular}
\end{table}
\par Figure \ref{fig:bb_aleph_designs} shows the noise amplification indices across all string routing designs that did not violate $\aleph_g(\mb{J}_{\ell c}) \geq 1\text{e-7}$, sorted in descending order of $\aleph_g\left( \mb{J}_{\ell\xi}(L) \right)$. Table \ref{table:bb_routing_aleph} shows the values $\aleph_g\left( \mb{J}_{\ell\xi}(L) \right)$, $\aleph_g\left( \mb{J}_{\ell\xi}(s_6) \right)$, and $\aleph_g\left( \mb{J}_{\ell\xi}(s_4) \right)$ for the designs that maximize each of these three values (which are also indicated with arrows in Fig. \ref{fig:bb_aleph_designs}). We observe that the design that maximizes $\aleph_g\left( \mb{J}_{\ell\xi}(s_6) \right)$ results in only a 0.8\% decrease in $\aleph_g\left( \mb{J}_{\ell\xi}(L) \right)$, however, the design that maximizes $\aleph_g\left( \mb{J}_{\ell\xi}(s_4) \right)$ results in a 90\% decrease in $\aleph_g\left( \mb{J}_{\ell\xi}(L) \right)$. Choosing the string routing design that maximizes $\aleph_g\left( \mb{J}_{\ell\xi}(s_4) \right)$ would therefore tend to increase the pose error at $s=L$. We also note that $\aleph_g\left( \mb{J}_{\ell\xi}(s_4) \right)$ only increases by 4.6\% between the end disk routing and the fourth disk routing, indicating that we would not expect to see a significant change in the pose error at $s_4$ between these two designs. We will now validate these predicted behaviors in a simulation study.
\begin{figure}[htbp]
  \centering
  \includegraphics[width=0.90\columnwidth]{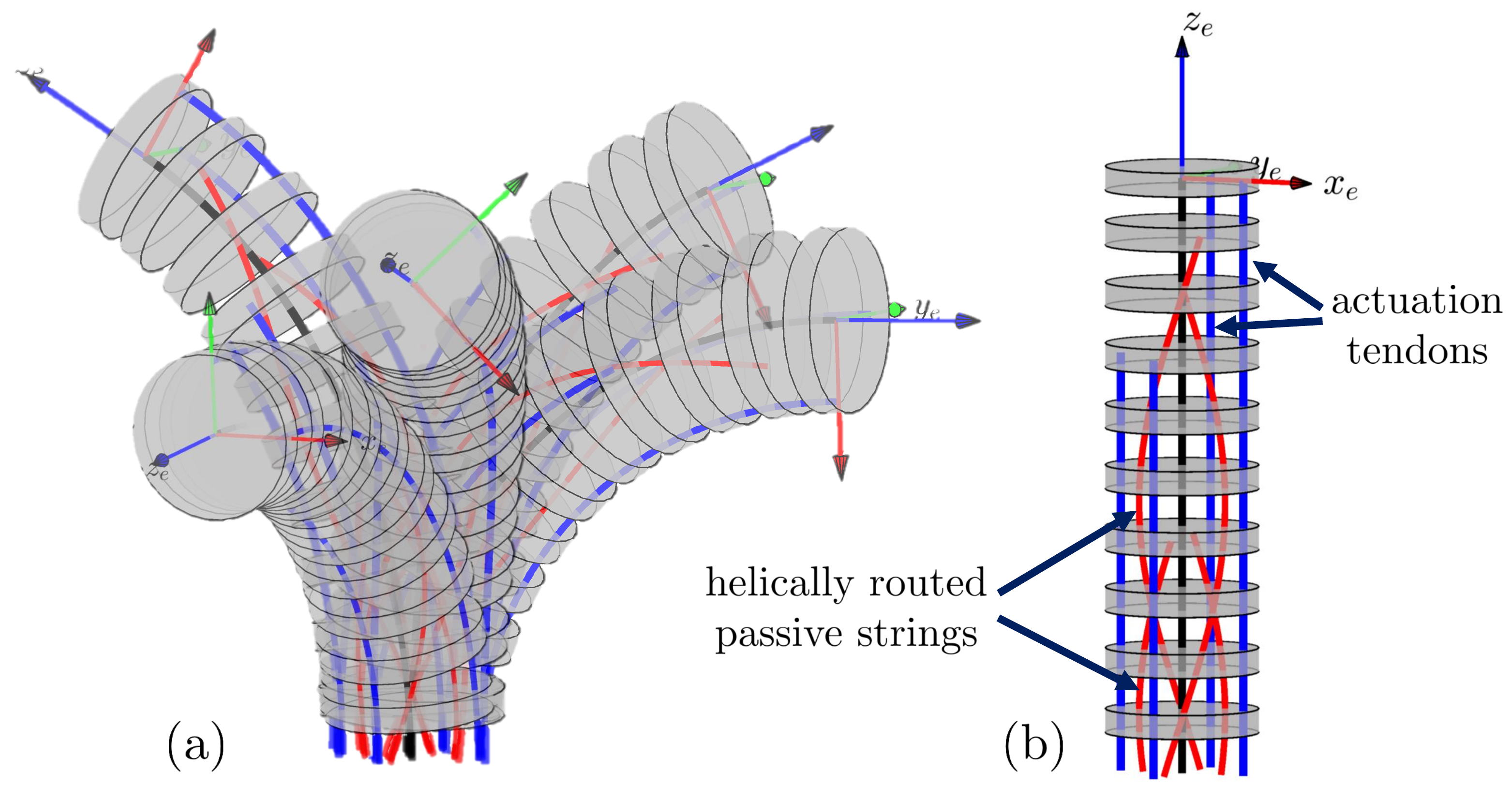}
  \caption{(a) A subset of the spatial configurations used to validate our shape sensing approach on a segment subject to torsional loads utilizing helical routing, and (b) the segment in its zero-curvature configuration.}
  \label{fig:bb_configs}
\end{figure}
\par We simulated the robot in Fig. \ref{fig:bandybot} using the Cosserat rod model from \cite{rucker2011statics}. This model takes as inputs the applied tensions on the actuation tendons as well as the external forces/moments applied to the tip of the segment and returns the curvature, shear, and extension of the segment. We simulated the robot with a preload force of 25 N applied on all four tendons, and applied additional forces of up to 300 N on the tendons to bend the segment in 8 different directions. For each direction, we applied 12 different external wrenches with forces of $\pm$20 N and moments of $\pm$ 4 Nm, expressed in the world frame. We selected these loads to generate a large variety of shapes without exceeding the maximum curvature to keep the segment within its admissible workspace. We also included constant forces of 3.3 N on the helical strings due to the constant-torque spring in the string encoder housing.
\par For each tendon tension/external wrench combination, we started with the segment initially unloaded and incremented the external wrench with 5 steps to incrementally apply the load and solve for the shape at each external wrench increment. We did this to ensure that a good initial guess was provided to the solver. This simulation resulted in 480 different configurations of the continuum segment. For each configuration, we stored the string/tendon lengths, and then used these lengths as inputs to our kinematic shape sensing model to compare the accuracy of our shape sensing approach. Our unoptimized MATLAB 2019b implementation used MATLAB's \emph{lsqnonlin()} to solve \eqref{eq:shape_sense_problem} for $\mb{c}$ at a rate of $\sim 4$ Hz on average, using an 80 point trapezoid rule to compute $\mb{J}_{\ell c}$. Future code can be orders of magnitude faster with direct implementation in C++.
\par We compared the mechanics model to two different shape sensing models. The first used all 8 length measurements (4 string encoders and 4 actuation tendons) with the modal basis given by \eqref{eq:bb_modal_funcs}. The second used only the 4 actuation tendons to compare our shape sensing model to a scenario without any helical shape sensing strings. For this second model, the modal basis has 4 columns and is given by \eqref{eq:curv_basis_zeros_torsion}, with the modal functions given by $\bs{\phi}_x(s) = \bs{\phi}_y(s) = \begin{bmatrix} T_0, T_1(s) \end{bmatrix}\T$.
\begin{table}[ht] \centering
\caption{Average (Maximal) Absolute Position and Orientation Errors in mm and $^\circ$ For the Segment in Fig.~\ref{fig:bandybot}. Errors Are Specified For the End Disk $(s=L)$ and the $4^{th}$ Disk $(s=s_4)$} \label{table:bb_errors}
\fontsize{8}{10}\selectfont \renewcommand {\arraystretch} {1.15}
\begin{tabular} {|p{26mm}|>{\centering\arraybackslash}p{13mm}|>{\centering\arraybackslash}p{14mm}|>{\centering\arraybackslash}p{14mm}|} \hline
                                                                   & End Disk Routing & Fourth Disk Routing & w/o Passive Strings \\ \thickhline
$\bar{p}_{e}(L)$, $\text{max}\left(p_{e}(L)\right)$                & 1.29 (3.39)      & 1.94 (5.64)        &  5.61 (32.87)   \\ \hline
$\bar{p}_{e}(s_4)$, $\text{max}\left(p_{e}(s_4)\right)$            & 0.45 (0.90)       & 0.52 (1.51)        &  0.63 (3.07)    \\ \hline
$\bar{\theta}_{e}(L)$, $\text{max}\left(\theta_{e}(L)\right)$      & 1.17 (3.04)      & 6.72 (21.78)       &  4.55 (23.44)   \\ \hline
$\bar{\theta}_{e}(s_4)$, $\text{max}\left(\theta_{e}(s_4)\right)$  & 0.61 (1.55)      & 0.57 (1.20)        &  2.15 (12.17)   \\\hline
\end{tabular}
\end{table}
\begin{figure}[htbp]
  \centering
  \includegraphics[width=0.90\columnwidth]{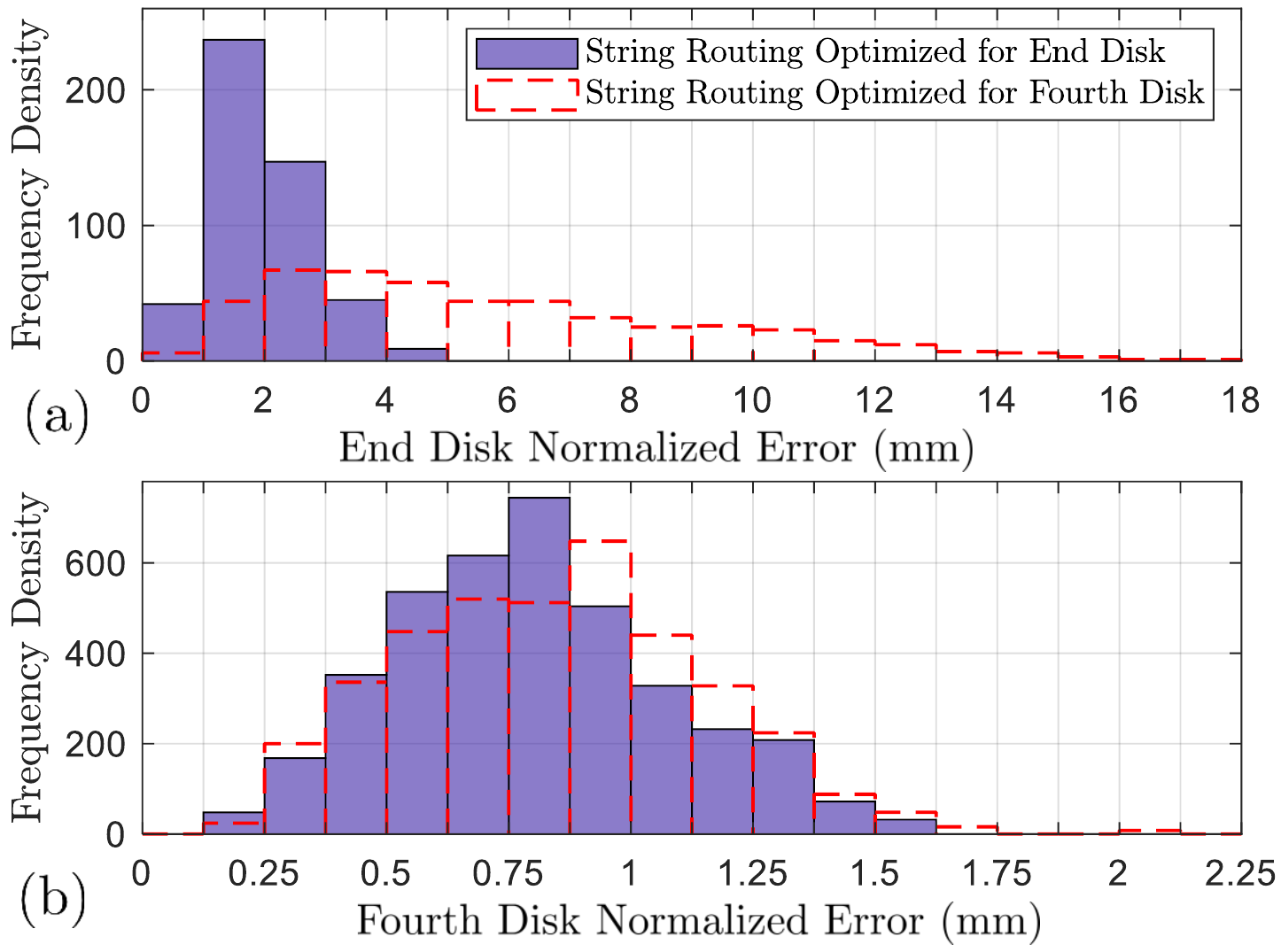}
  \caption{ Histograms of normalized pose error at the end disk and at the fourth intermediate disk for the routing that maximizes $\aleph_g(\mb{J}_{\ell \xi}(L))$ and for the routing that maximizes $\aleph_g(\mb{J}_{\ell \xi}(s_4))$. The pose error at the fourth intermediate disk is not significantly effected by the change in routing design, but the tip pose error is significantly effected due to the larger change in $\aleph_g(\mb{J}_{\ell \xi}(L))$ between the two designs. }
  \label{fi:bb_histo}
\end{figure}

\par The statistical results are given in Table \ref{table:bb_errors}, where we denote the string routing design that maximizes $\aleph_g\left( \mb{J}_{\ell\xi}(L) \right)$ as the \emph{end disk routing} and the string routing design that maximizes $\aleph_g\left( \mb{J}_{\ell\xi}(s_4) \right)$ as the \emph{fourth disk routing}. Both optimized string routing designs had maximum absolute end disk position errors below 2\% of arc length. However, the fourth disk routing resulted in a 616\% increase in $\text{max}(\theta_e(L))$ compared to the end disk routing. There was also a 474\% increase in $\bar{\theta}_{e}(L)$, and small increases in both $\bar{p}_{e}(L)$ and $\text{max}\left(p_{e}(L)\right)$ when compared to the fourth disk routing. This confirms the expected behavior due to the large decrease in $\aleph_g\left( \mb{J}_{\ell\xi}(L) \right)$ in the fourth disk routing given in Table \ref{table:bb_routing_aleph}.
\par Figure \ref{fi:bb_histo} shows the histograms of the error at the end disk and the fourth disk, normalized using \eqref{eq:normalized_error}. We observe that the end disk error with the end disk routing is shifted leftward compared to the fourth disk routing errors, as expected due to the large change in $\aleph_g\left( \mb{J}_{\ell\xi}(L) \right)$. Furthermore, since $\aleph_g\left( \mb{J}_{\ell\xi}(s_4) \right)$ did not significantly change, we do not see a significant difference in the error distribution for the fourth disk pose, and in fact see a small shift rightward with the fourth disk routing. This increase in error using the fourth disk routing is explained by the known fact that kinematic conditioning indices are not guaranteed to directly correlate with the true errors (see \cite{merlet2005jacobian}). Large changes in the conditioning index should be sought to increase the possibility of reducing the true errors.
\par Compared to the model without string measurements, the end disk routing reduced the maximum end disk position error by 90\% and the maximum angular end disk error by 87\%. The fourth disk routing reduced the fourth disk maximum position error by 51\% and the fourth disk maximum angular error by 90\%. These simulation results demonstrate that 4 passive string (together with the four actuation tendons) can significantly improve the accuracy of continuum robot kinematics over actuation-based sensing alone.
\section{Conclusions}
\par In this paper, we have presented a Lie group kinematic formulation for capturing variable curvature deflections of continuum robots using general string encoder routing. We used this formulation for a sensitivity analysis of the error propagation from error in string extension measurements to error in the modal coefficients and error in the central backbone shape. This analysis allows the designer to avoid string encoder routings that lead to ill-conditioned Jacobians. We then applied the approach on a planar example, a segment with high torsional stiffness, and a robot subject to torsional deflections using helical routing, showing that this shape sensing approach can result in mean and maximal absolute position error below 2\% and 5\% of arc length, respectively.
\par Our results provided several simple design guidelines for routing the strings to improve numerical conditioning, which we summarize here. For a planar segment, the strings should not be anchored at the same disk, and the pitch radius should be maximized. For a segment with high torsional stiffness, no more than two strings should be anchored to a disk, and if two strings are anchored to a disk, their anchor points should not be collinear in the radial direction of the disk. For segments with general deflections including torsion, simple design rules are more difficult to define, but the numerical conditioning can be improved using our proposed design optimization procedure.
\par This sensing approach utilizes standard mechanical and electrical components, providing a relatively low-cost way of sensing the variable curvature deflections of large continuum robots, and the kinematic formulation and analysis presented herein enables practitioners to design string routings that improve the numerical conditioning of shape sensing. We have shown that four string encoders can provide accurate shape sensing, and have demonstrated a physical embodiment of a segment utilizing this approach, but a drawback of this approach compared to other sensing methods is the physical space required to mount the string encoders within the robot. To overcome this drawback, directions for future work include investigating more tightly integrated design of the string encoder mechanical components into the robot body or mounting of the string encoder housing remotely outside of the robot body. Future work also includes evaluation on multi-segment continuum robots and using this shape sensing approach and Lie group kinematic formulation to provide updates to continuum robot mechanics models.
\section*{Acknowledgements}
Thank you to Zachary Taylor for help with the code to read and calibrate the string encoders.
\bibliographystyle{IEEEtran}
\bibliography{IEEEabrv,shape_sensing}
%
%
\begin{IEEEbiography}[{\includegraphics[width=1in,clip,keepaspectratio]{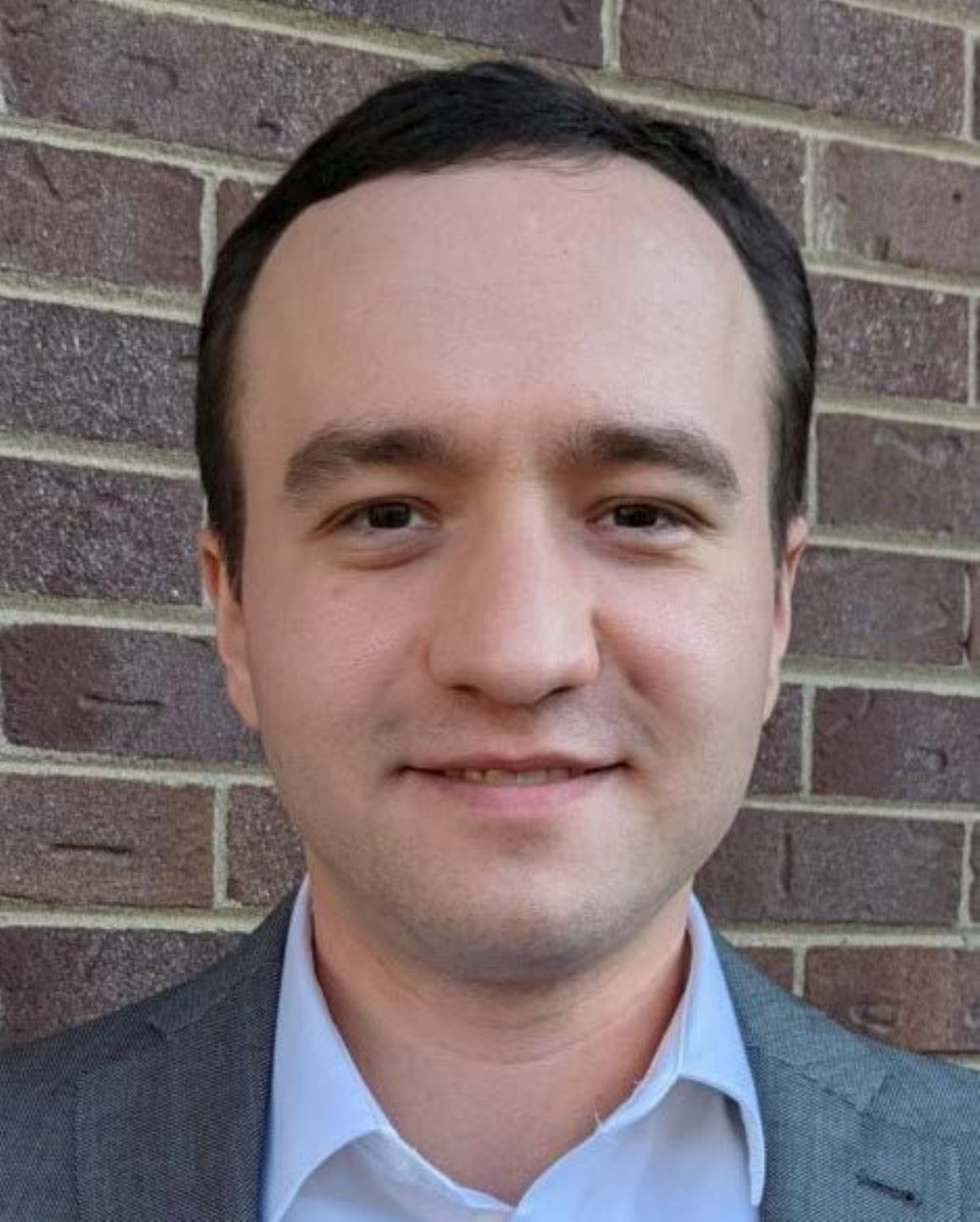}}]{Andrew L. Orekhov} received the B.S. degree in mechanical engineering from the University of Tennessee, Knoxville, TN, in 2016, and the Ph.D. in mechanical engineering from Vanderbilt University, Nashville, TN, in 2022. He is currently a Postdoctoral Fellow in the Robotics Institute at Carnegie Mellon University and his research interests include design, modeling, and control of continuum manipulators and snake robots. He received the NSF Graduate Research Fellowship in 2016.
\end{IEEEbiography}
\begin{IEEEbiography}[{\includegraphics[width=1in,clip,keepaspectratio]{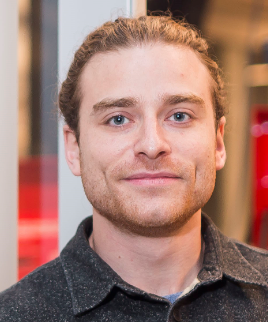}}]{Elan Z. Ahronovich} received his B.S. degree in biological sciences from the Ohio State University, Columbus, OH, USA, in 2014 and his B.S. degree in mechanical engineering from Virginia Polytechnic Institute and State University, Blacksburg, VA, USA, in 2019. His current research interests include collaborative surgical robotics and soft continuum manipulators.
\end{IEEEbiography}
\begin{IEEEbiography}[{\includegraphics[width=1in,clip,keepaspectratio]{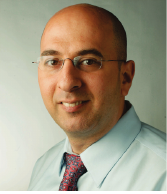}}]{Nabil Simaan} (F’20) received his Ph.D. degree in mechanical engineering from the Technion—Israel Institute of Technology, Haifa, Israel, in 2002. During 2003, he was a Postdoctoral Research Scientist at Johns Hopkins University National Science Foundation (NSF) ERC-CISST. In 2005, he joined Columbia University, New York, NY. In 2009 he received the NSF Career award for young investigators to design new algorithms and robots for safe interaction with the anatomy. In Fall 2010 he joined Vanderbilt University. In 2020 he was named an IEEE Fellow for  contributions to dexterous continuum robotics. In 2021 he was elected Fellow of the ASME for contributions to continuum and soft robotics for surgery. His research interests include medical robotics, kinematics, robot modeling and control and human-robot interaction.
\end{IEEEbiography}
\balance
\end{document}